\documentclass[pdflatex,sn-mathphys-num]{sn-jnl}

\usepackage{graphicx}%
\usepackage{multirow}%
\usepackage{amsmath,amssymb,amsfonts}%
\usepackage{amsthm}%
\usepackage{mathrsfs}%
\usepackage[title]{appendix}%
\usepackage{textcomp}%
\usepackage{manyfoot}%
\usepackage{algorithm}%
\usepackage{algorithmicx}%
\usepackage{algpseudocode}%
\usepackage{listings}%

\theoremstyle{thmstyleone}%
\theoremstyle{thmstyletwo}%

\theoremstyle{thmstylethree}%

\newcommand{\annotator}[2]{\csdef{#1}##1{{\color{#2} [\textbf{\MakeUppercase #1}: ##1]}}}
\annotator{liang}{cyan}
\annotator{roy}{green}

\usepackage{xspace}
\newcommand{\themodel}{DiffSpectra\xspace}

\usepackage[svgnames,dvipsnames,table]{xcolor}

\usepackage{todonotes}

\usepackage[capitalize]{cleveref}
\crefname{figure}{Figure}{Figures}
\crefname{table}{Table}{Tables}
\crefformat{equation}{Eq.~(#2#1#3)}

\usepackage{geometry}
\usepackage{caption}
\usepackage{my_math}
\usepackage{booktabs}
\usepackage{threeparttable}
\usepackage{colortbl}
\usepackage{multirow}
\usepackage{makecell}
\usepackage{float}
\usepackage[version=4]{mhchem}

\usepackage{pifont}
\newcommand{\cmark}{\ding{51}}
\newcommand{\xmark}{\ding{55}}

\usepackage{array}
\usepackage{ragged2e}
\newcolumntype{P}[1]{>{\RaggedRight\hspace{0pt}}p{#1}}

\newcommand{\first}[1]{\textbf{#1}}

\begin{document}

\title[DiffSpectra]{\themodel: Molecular Structure Elucidation from Spectra using Diffusion Models}


\author[1,2,6]{\fnm{Liang} \sur{Wang}}\email{liang.wang@cripac.ia.ac.cn}
\author*[3,4]{\fnm{Yu} \sur{Rong}}\email{yu.rong@hotmail.com}
\author[3,4]{\fnm{Tingyang} \sur{Xu}}\email{xuty\_007@hotmail.com}
\author[5]{\fnm{Zhenyi} \sur{Zhong}}\email{zhenyi\_zhong@tju.edu.cn}
\author[6]{\fnm{Zhiyuan} \sur{Liu}}\email{acharkq@gmail.com}
\author[3,4]{\fnm{Pengju} \sur{Wang}}\email{weichang.wpj@alibaba-inc.com}
\author[3,4]{\fnm{Deli} \sur{Zhao}}\email{zhaodeli@gmail.com}
\author*[1,2]{\fnm{Qiang} \sur{Liu}}\email{qiang.liu@nlpr.ia.ac.cn}
\author[1,2]{\fnm{Shu} \sur{Wu}}\email{shu.wu@nlpr.ia.ac.cn}
\author*[1,2]{\fnm{Liang} \sur{Wang}}\email{wangliang@nlpr.ia.ac.cn}
\author*[6,7,8]{\fnm{Yang} \sur{Zhang}}\email{zhang@zhanggroup.org}

\affil[1]{\orgdiv{NLPR, MAIS, Institute of Automation}, \orgname{Chinese Academy of Sciences}, \orgaddress{\state{Beijing}, \country{China}}}
\affil[2]{\orgdiv{School of Artificial Intelligence}, \orgname{University of Chinese Academy of Sciences},\\ \orgaddress{\state{Beijing}, \country{China}}}
\affil[3]{\orgdiv{DAMO Academy}, \orgname{Alibaba Group}, \orgaddress{\state{Hangzhou}, \country{China}}}
\affil[4]{\orgname{Hupan Lab}, \orgaddress{\state{Hangzhou}, \country{China}}}
\affil[5]{\orgdiv{College of Intelligence and Computing}, \orgname{Tianjin University}, \orgaddress{\state{Tianjin}, \country{China}}}
\affil[6]{\orgdiv{Department of Computer Science, School of Computing}, \orgname{National University of Singapore}, \orgaddress{\state{Singapore}, \country{Singapore}}}
\affil[7]{\orgdiv{Department of Biochemistry, Yong Loo Lin School of Medicine}, \orgname{National University of Singapore}, \orgaddress{\state{Singapore}, \country{Singapore}}}
\affil[8]{\orgdiv{Cancer Science Institute of Singapore}, \orgname{National University of Singapore}, \orgaddress{\state{Singapore}, \country{Singapore}}}


\abstract{
Molecular structure elucidation from spectra is a fundamental challenge in molecular science.
Conventional approaches rely heavily on expert interpretation and lack scalability, while retrieval-based machine learning approaches remain constrained by limited reference libraries. Generative models offer a promising alternative, yet most adopt autoregressive architectures that overlook 3D geometry and struggle to integrate diverse spectral modalities.
In this work, we present DiffSpectra, a generative framework that formulates molecular structure elucidation as a conditional generation process, directly inferring 2D and 3D molecular structures from multi-modal spectra using diffusion models.
Its denoising network is parameterized by the Diffusion Molecule Transformer, an SE(3)-equivariant architecture for geometric modeling, conditioned by SpecFormer, a Transformer-based spectral encoder capturing multi-modal spectral dependencies.
Extensive experiments demonstrate that DiffSpectra accurately elucidates molecular structures, achieving 40.76\% top-1 and 99.49\% top-10 accuracy. Its performance benefits substantially from 3D geometric modeling, SpecFormer pre-training, and multi-modal conditioning.
To our knowledge, DiffSpectra is the first framework that unifies multi-modal spectral reasoning and joint 2D/3D generative modeling for de novo molecular structure elucidation.
}




\maketitle

\section{Introduction}\label{sec:introduction}

Molecular structure elucidation represents one of the fundamental challenges in chemistry, materials science, and biology, underpinning compound identification, reaction mechanism elucidation, and drug discovery~\cite{MolDiscovery,multimodal-spec,MaterialSurvey}. 
This process involves the accurate determination of two-dimensional (2D) atomic connectivity and three-dimensional (3D) molecular conformation based on experimental data.
Traditional approaches often rely on expert interpretation and substantial manual effort to propose and validate structural hypotheses. Although these methods have proven invaluable over decades of chemical research, the growing complexity of natural products, synthetic compounds, and pharmaceutical intermediates calls for more sophisticated computational approaches that can efficiently navigate the vast chemical space of possible structures.

The integration of machine learning into spectroscopic analysis has opened new avenues for automating and accelerating molecular structure elucidation. 
Early applications of machine learning in molecular structure elucidation primarily adopted retrieval-based paradigms~\cite{MolDiscovery,Spectra2Structure,MassSpecGym}. These approaches treat the task as a matching problem: given an experimental spectrum, the model searches a predefined molecular library to retrieve candidate structures whose spectra best align with that spectrum. Although efficient in practice, these methods are inherently constrained by the coverage and quality of the reference library. To mitigate this limitation, some efforts have proposed spectral prediction models, which simulate spectra from known molecular structures~\cite{NEIMS,PaiNN,DetaNet,MassFormer,SCARF,MoMS-Net,DeepGP}. These predicted spectra are then used to expand the searchable library and improve retrieval coverage. However, such methods still rely on the existence of a finite reference library and thus fundamentally struggle to generalize to novel molecules outside it, thereby limiting their applicability in open-ended or exploratory chemical analyzes.

To overcome reliance on predefined molecular libraries and advance toward genuinely data-driven elucidation, recent efforts have turned to more expressive neural architectures capable of inferring  molecular structures \textit{de novo} from spectral measurements.
Initial attempts in this direction have framed the task as a predictive problem, training models to infer coarse structural features from spectra, such as molecular properties~\cite{Spectra2Property}, fingerprints~\cite{MIST}, fragments~\cite{LC-MS,NMR2Struct}, chemical formulas~\cite{FIDDLE}, or other approximate structural profiles~\cite{DeepMSProfiler,MALDI-MSI-IMC,OrbitrapAstral,CrystalNet}. Although valuable, these coarse predictions fall short of achieving a complete molecular structure elucidation.

In a shift toward complete structure elucidation, subsequent works have introduced generative models that take spectral data as input and produce discrete molecular representations, such as SMILES strings or molecular graphs. These models typically follow autoregressive architectures, generating molecular structures token-by-token using RNNs~\cite{Spec2Mol,Seq2Seq,NMRGen,MSNovelist} or Transformers~\cite{Casanovo,NMR2Struct,Spectra2Structure,IR2Structure,Vib2Mol,PatchBasedSelfAttention}. Despite their promise, these models face two fundamental limitations. First, generative models built on discrete SMILES strings or graphs inherently lack geometric inductive bias and cannot effectively capture 3D geometric information. This is critical because many spectral modalities, such as infrared (IR), Raman, and ultraviolet-visible (UV-Vis), are inherently dependent on the geometry of molecules and the associated energy states~\cite{MolSpectra}. 
Second, all existing methods are limited to processing a single type of spectrum in isolation, whereas molecular structure elucidation, in practice, relies on the joint interpretation of multiple spectral modalities, each offering complementary structural cues~\cite{MolPuzzle}. Current models lack support for such multi-modal spectral reasoning, thereby constraining their accuracy and generalizability. These limitations highlight the need for a principled,  unified generative framework that can integrate both geometric reasoning and multi-modal spectral conditioning.

The inherent limitations of existing approaches motivate the development of more sophisticated generative models that can effectively bridge the gap between spectral observations and molecular structure determination. 
Diffusion models have recently attracted significant attention in generative modeling due to their ability to produce high-quality samples through iterative denoising processes~\cite{Diffusion,DDPM,SMLD,SDE,VDM}. 
This framework has demonstrated remarkable success in modeling complex, multi-modal distributions, generating diverse outputs, and showing particular promise in molecular applications~\cite{MolecularDiffusionSurvey,MaterialSurvey,EDM,GeoLDM,DiffLinker,DiffSBDD,GaUDI,PMDM}.
The ability of diffusion models to generate chemically meaningful structures and handle multi-modal distributions makes them particularly well-suited for molecular structure elucidation.

In this work, we introduce \themodel, a framework that leverages diffusion models to directly generate molecular structures from multi-modal spectra. Our approach addresses the structure elucidation problem as a conditional generation task, where spectral measurements serve as conditioning information that guides the iterative denoising process toward generating plausible molecular structures.

The core of our framework is a joint 2D and 3D diffusion model that simultaneously generates molecular topological structures and geometric coordinates. Each molecule is represented as a geometric graph with node features (e.g., atom types), edge features (e.g., bond types), and 3D coordinates, thereby capturing both topological structure and geometric conformation in a unified representation. \themodel formulates structure elucidation as a continuous-time conditional denoising process, where noise is progressively removed under spectral constraints. By jointly modeling topological and geometric components, the model ensures that the generated structures satisfy chemical bonding rules while preserving geometric consistency. This dual-space generation is essential for accurately elucidating molecules whose spectra reflect both bonding connectivities and 3D conformations, effectively leveraging the complementary strengths of topology and geometry to produce structurally consistent outputs.

To parameterize the denoising process, we introduce the Diffusion Molecule Transformer (DMT), an SE(3)-equivariant architecture designed for molecular structure generation. DMT comprises a stack of transformer-style blocks operating on three parallel streams of node features, edge features, and coordinates, enabling rich interactions between topological and geometric representations. Through relational multi-head attention and equivariant geometric updates, DMT effectively captures spatial relationships while preserving chemical symmetry.

To effectively condition the generation process on multi-modal spectra, we introduce SpecFormer, a Transformer-based spectrum encoder specifically designed to process and integrate multiple types of molecular spectra simultaneously, including IR, Raman, and UV-Vis spectra. SpecFormer partitions each spectrum into local patches, embeds them into a unified latent space, and employs multi-head self-attention to capture both intra- and inter-spectrum dependencies. 
To enhance the quality of spectral embeddings, we pre-train SpecFormer using masked patch reconstruction and contrastive alignment with molecular structures.
The resulting spectral embeddings provide conditioning signals for \themodel, effectively guiding the generation toward structures consistent with the measured spectra.

We extensively evaluated \themodel to assess its effectiveness in elucidating molecular structures from multi-modal spectra. Our results show that \themodel generates chemically valid and stable molecular structures. When conditioned on spectra, our model achieves a top-1 exact-structure accuracy of 40.8\%, and reconstructs key functional groups with over 96\% similarity. We further show that using a pre-trained SpecFormer significantly improves structure-elucidation accuracy, and that combining multiple spectral modalities yields better performance than any single spectrum alone. Importantly, we find that sampling multiple candidates greatly enhances the hit rate: the top-10 accuracy rises to 99.49\%, indicating that correct structures are consistently ranked among the plausible outputs. This highlights the practical value of diffusion-based sampling in structure elucidation, as generating a shortlist of candidates enables reliable downstream validation. Our ablations further confirm the advantage of generating 3D geometries and incorporating SE(3)-equivariance, and demonstrate that moderate sampling temperatures strike a good balance between diversity and accuracy.
Further, \themodel demonstrates robust generalization across molecules with varying numbers of atoms and exhibits an elucidation trajectory in which simpler substructures are resolved first, followed by more complex ones, as revealed by gradient analysis.
These findings highlight the potential of spectra-guided diffusion models for accurate molecular structure elucidation.

To the best of our knowledge, this is the first work to leverage diffusion models for elucidating complete 3D molecular structures directly from multi-modal spectral data.
\section{Results}\label{sec:results}

\subsection{Overview of DiffSpectra Framework}

\begin{figure}[H]
\begin{center}
\includegraphics[width=1\linewidth]{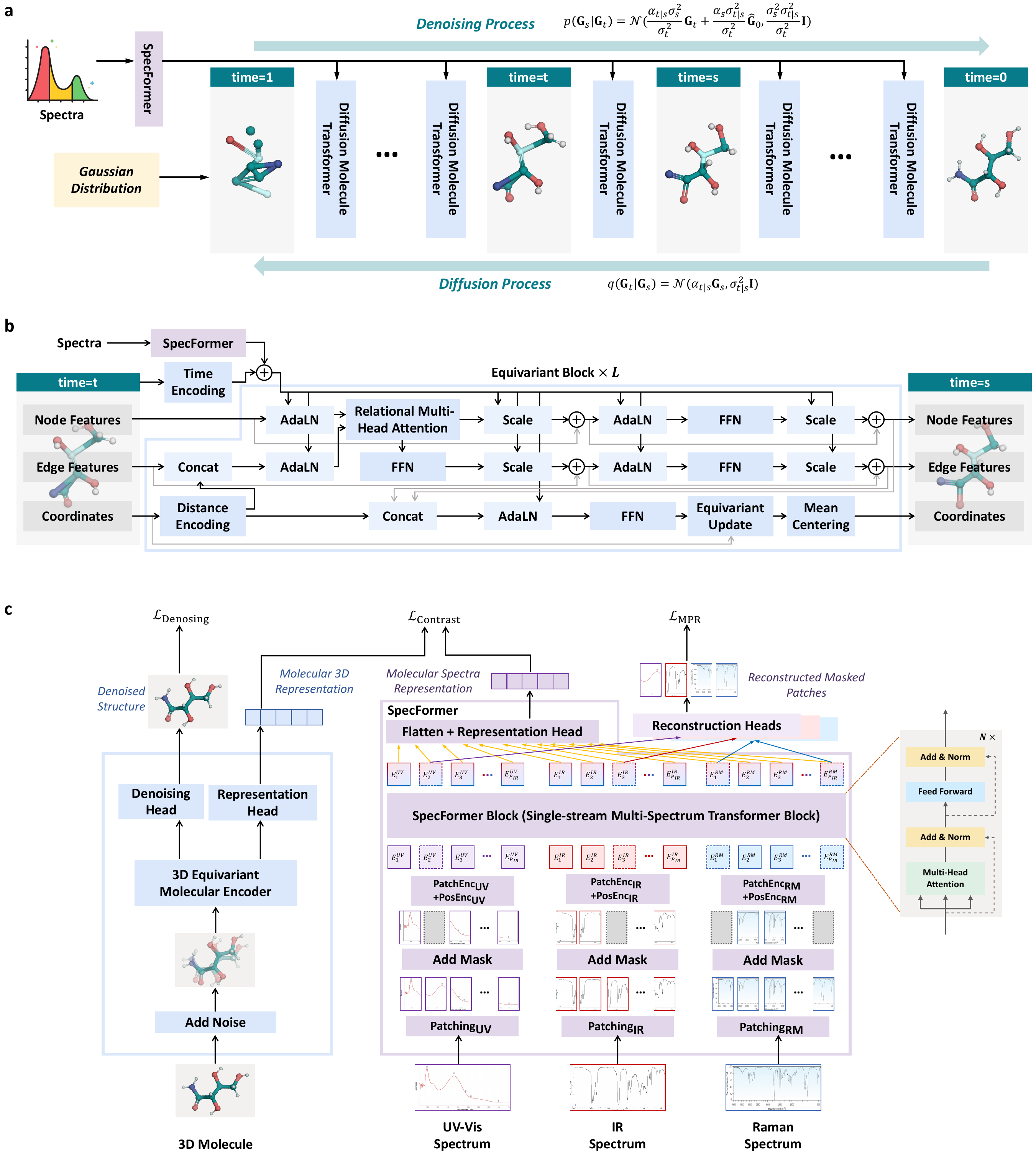}
\end{center}
\caption{\textbf{Overview and architecture of the \themodel framework.} \textbf{a,} Schematic of the diffusion-based generative framework underlying \themodel, comprising a continuous-time forward diffusion process and a reverse-time denoising process. The denoising network is implemented as the Diffusion Molecule Transformer (DMT), while spectral features encoded by SpecFormer provide conditional guidance along the denoising trajectory.
\textbf{b,} Architecture of DMT, which jointly processes and denoises node features, edge features, and atomic coordinates through three parallel streams within a stack of equivariant blocks. The streams are interconnected by a relational multi-head attention mechanism and equivariant geometric updates, and are conditioned on shared spectral and time information.
\textbf{c,} Architecture and pre-training strategy of SpecFormer, a unified Transformer encoder for multi-modal spectroscopic data (UV–Vis, IR, and Raman). SpecFormer is pre-trained with masked-patch reconstruction (MPR) and contrastive learning against corresponding 3D molecular structures.}
\label{fig:overview}
\end{figure}

To achieve accurate elucidation of molecular structure from molecular spectra, we propose \themodel, a spectra-conditioned diffusion framework that generates molecular structures guided by spectral information. The overall framework is illustrated in \cref{fig:overview}. 

Formally, a molecule characterized by its structure and spectra is represented as $\mathcal{M} = (\mathcal{G}, \mathcal{S})$. The molecular structure is modeled as a graph $\mathcal{G} = (\mathbf{H}, \mathbf{A}, \mathbf{X})$, where $\mathbf{H} \in \mathbb{R}^{N \times d_1}$ denotes node-level attributes (e.g., atom types and charges), $\mathbf{A} \in \mathbb{R}^{N \times N \times d_2}$ encodes pairwise edge features (e.g., bond existence and bond types), and $\mathbf{X} \in \mathbb{R}^{N \times 3}$ specifies the 3D coordinates of the atoms. Here, $N$ is the number of atoms, $d_1$ and $d_2$ are the node and edge feature dimensions. Atomic ordering is kept consistent across $\mathbf{H}$ and $\mathbf{X}$ to preserve the correspondence between atom types and spatial positions. 

Meanwhile, the molecular spectra are denoted as $\mathcal{S} = (\vs_1, \ldots, \vs_{|\mathcal{S}|})$, where $|\mathcal{S}|=3$ in our setting, corresponding to UV-Vis, IR, and Raman spectra. Specifically, $\vs_1 \in \mathbb{R}^{601}$ represents the UV-Vis spectrum from 1.5 to 13.5 eV with 0.02 eV intervals; $\vs_2 \in \mathbb{R}^{3501}$ denotes the IR spectrum spanning 500–4000 cm$^{-1}$ at 1 cm$^{-1}$ resolution; and $\vs_3 \in \mathbb{R}^{3501}$ denotes the Raman spectrum on the same range. Together, these modalities provide a comprehensive description of molecular signatures.

Within DiffSpectra, a continuous-time variational diffusion model progressively perturbs the molecular graph with Gaussian noise during the forward diffusion process, and then denoises it in reverse, conditioned on spectra-derived information. The denoising process is parameterized by the Diffusion Molecule Transformer, which preserves permutation and SE(3) equivariance. DMT is guided by spectral embeddings extracted by SpecFormer, a Transformer-based encoder pre-trained with both masked reconstruction and contrastive alignment to capture spectrum–structure correlations. This design enables faithful and chemically meaningful structure elucidation, supported by physically grounded diffusion and rich spectral priors. More details of the full architecture and training objectives are provided in \cref{sec:method}.

\begin{table*}[h]
\centering
\caption{
Evaluation of basic molecular generation metrics on the QM9S dataset.
Metrics include stability, validity, uniqueness, novelty, and distribution-based metrics. 
The \textit{Training Set} row reports statistics of the training set as a reference distribution. 
\themodel is compared against unconditional molecular generation models.
}
\label{tab:validity}
\scriptsize
\begin{minipage}{\textwidth}
\centering
\makebox[\textwidth][c]{
\resizebox{1.1\textwidth}{!}{
\begin{tabular}{cc|cccccccc}
\toprule
\multicolumn{2}{c|}{\textbf{\textit{Models}}} & AtomStable $\uparrow$ & MolStable $\uparrow$ & V\&U $\uparrow$ & V\&U\&N $\uparrow$ & FCD $\downarrow$ & SNN $\uparrow$ & Frag $\uparrow$ & Scaf $\uparrow$ \\
\midrule
\multicolumn{2}{c|}{\textit{Training Set}} & \textit{99.9\%} & \textit{98.8\%} & \textit{98.9\%} & \textit{0.0\%} & \textit{0.063} & \textit{0.490} & \textit{0.992} & \textit{0.946} \\
\midrule
\multirow{2}{*}{\footnotesize{\makecell[c]{Unconditional Molecular\\ Generation}}} & CDGS~\cite{CDGS} & 99.7\% & 95.1\% & 93.6\% & 89.8\% & 0.798 & 0.493 & 0.973 & 0.784 \\
~ & JODO~\cite{JODO} & \textbf{99.9\%} & \textbf{98.8\%} & {96.0\%} & 89.5\% & {0.138} & {0.522} & {0.986} & {0.934} \\
\midrule
\footnotesize{\makecell[c]{Molecular Structure\\ Elucidation}} & \themodel & \textbf{99.9\%} & 98.2\% & \textbf{96.8\%} & \textbf{93.7\%} & \textbf{0.088} & \textbf{0.531} & \textbf{0.992} & \textbf{0.943} \\
\bottomrule
\end{tabular}
}
}
\end{minipage}
\end{table*}

\subsection{\themodel generates valid and stable molecular structures}

As summarized in \cref{tab:validity}, our proposed \themodel consistently achieves strong performance in generating chemically valid and stable molecular structures under spectra-conditioned settings. It attains 99.9\% atom stability and 98.2\% molecular stability, demonstrating performance comparable to that of the training data distribution, and outperforming or matching general unconditional molecular generation models such as CDGS~\cite{CDGS} and JODO~\cite{JODO}. \themodel also maintains high structural uniqueness and novelty, with V\&U (valid and unique) at 96.8\% and V\&U\&N (valid, unique, and novel) at 93.7\%, significantly surpassing prior methods. On distribution-based metrics, \themodel achieves the lowest Fréchet ChemNet Distance (FCD), indicating a close alignment with the test distribution, and the best SNN, Frag, and Scaf scores, demonstrating strong coverage of diverse chemical substructures and scaffolds while maintaining validity.

The \textit{Training Set} row in \cref{tab:validity} reports the statistics of the training set as the reference distribution. For metrics that measure similarity to the test set, such as SNN, Frag, or Scaf, the training data do not necessarily achieve the highest scores, since the test distribution may differ from the training distribution. All other baselines listed in \cref{tab:validity} are unconditional molecular generation models that sample the general molecular space without explicit conditions. In contrast, \themodel is specifically designed for molecular-structure elucidation conditioned on spectra. Spectral signals provide external structural priors, enabling the model to generate structures that better match the test set distribution. Therefore, the higher V\&U, V\&U\&N, FCD, SNN, Frag, and Scaf metrics achieved by \themodel compared to unconditional models are expected, as the conditioning naturally constrains the generation process toward structures consistent with the observed test samples.

\subsection{\themodel accurately elucidates molecular structures from spectra}

\begin{figure}[H]
\begin{center}
\includegraphics[width=\linewidth]{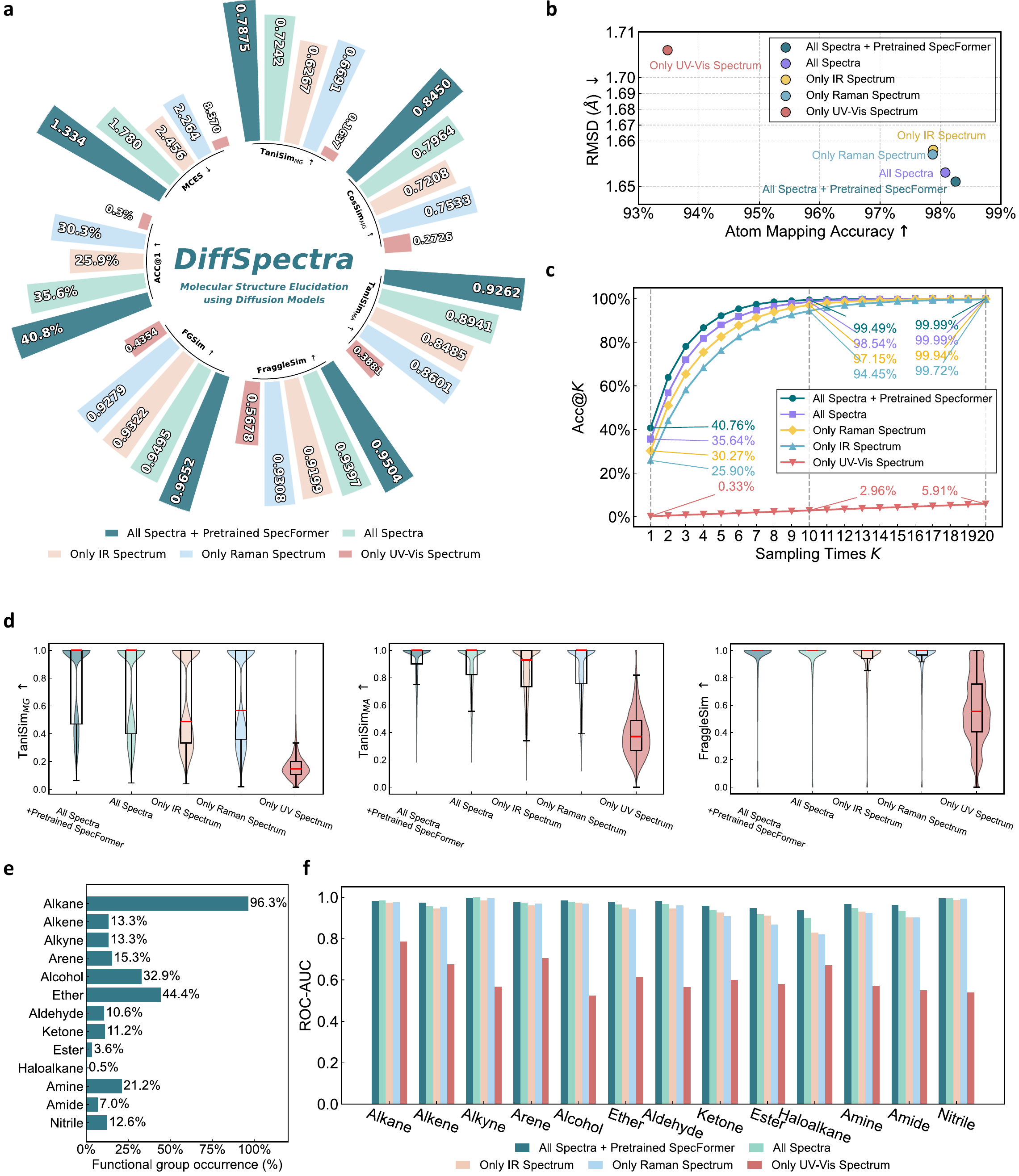}
\end{center}
\caption{\textbf{Quantitative comparison of \themodel and its variants on molecular structure elucidation tasks.} 
\textbf{a}, Structure elucidation performance of DiffSpectra under different configurations. We compared the performance when using a pre-trained SpecFormer as the spectral condition encoder versus an un-pretrained SpecFormer, as well as when using multi-modal spectra versus individual spectra (IR, Raman, UV-Vis). Reported metrics include exact top-1 accuracy, MCES, and several similarity-based metrics.
\textbf{b}, Evaluation of elucidated 3D molecular structures. For each generated 3D structure, we performed atom-to-atom mapping against the corresponding ground-truth 3D structure and further computed the RMSD. When using a pre-trained SpecFormer with multi-modal spectra as input, DiffSpectra achieves the best elucidation performance, showing the highest atom mapping accuracy and the lowest RMSD.
\textbf{c}, Accuracy@$K$ with an increasing number of sampled candidates. We report top-$K$ accuracy as the number of generated candidates $K$ increases. Across all settings, Accuracy@$K$ consistently improves with larger $K$, confirming that multiple sampling significantly increases the likelihood of recovering the correct molecular structure. 
\textbf{d}, Distribution of similarity-based metrics on the test set, illustrated using violin and box plots. The red line in each box plot denotes the median.
\textbf{e} and \textbf{f}, Occurrence of functional groups in the test molecules and the corresponding elucidation performance of DiffSpectra. To account for the imbalance in functional group occurrence, we computed the ROC-AUC score for each group based on whether DiffSpectra correctly identified its presence or absence in each molecule.
}
\label{fig:performance}
\end{figure}

\begin{figure}[t]
\begin{center}
\includegraphics[width=\linewidth]{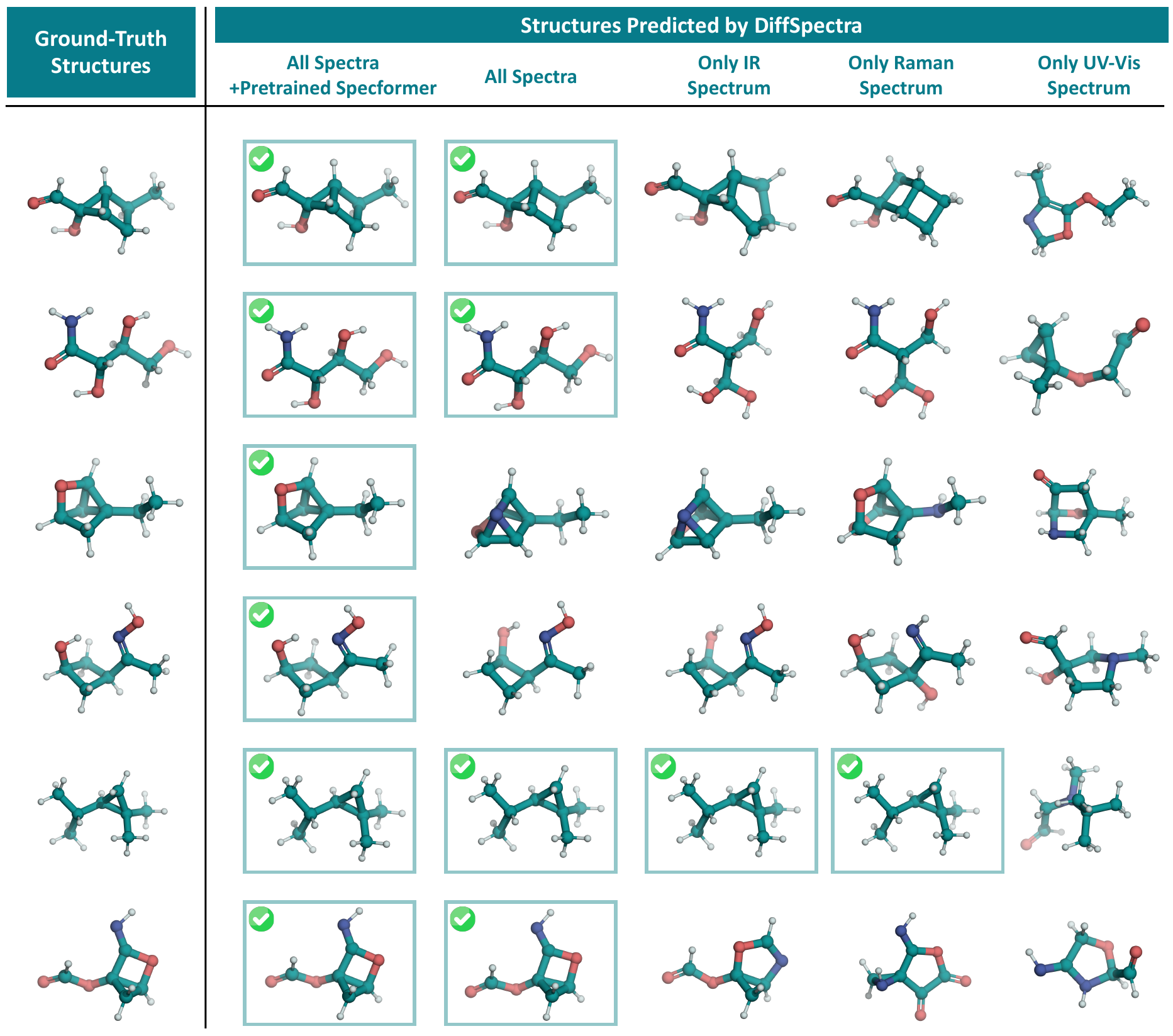}
\end{center}
\caption{Visualization of molecular structure elucidation results using DiffSpectra under different configurations. We compared single-spectrum inputs (IR, Raman, UV-Vis), multi-modal spectra, and the effect of the pre-trained SpecFormer. Ground-truth structures are shown on the left for reference.}
\label{fig:visualization}
\end{figure}

As shown in \cref{fig:performance}(A), \cref{tab:specformer}, and \cref{fig:visualization}, \themodel demonstrates strong performance on the challenging task of molecular structure elucidation from spectra. The top-1 accuracy (Acc@1) reaches 40.76\%, indicating that the model can recover the exact target structure among its predictions in a non-trivial fraction of cases. Beyond strict exact matches, the average MCES score is 1.3337, suggesting that even when the model fails to generate an exact structure, it still preserves a substantial portion of the molecular graph connectivity.

In terms of fingerprint-based similarity metrics, \themodel achieves a Tanimoto similarity of 0.7875 and a cosine similarity of 0.8450 over Morgan fingerprints, reflecting high agreement with local structural features. The Tanimoto similarity over MACCS keys is even higher at 0.9262, highlighting broad coverage of predefined chemical substructures. These results suggest that \themodel effectively captures the key functional groups and local motifs characteristic of the target molecules.
Furthermore, the fragment-based Fraggle similarity reaches 0.9504, showing that large chemically meaningful fragments are well recovered. The functional group similarity (FGSim) is also high at 0.9652, confirming that the predicted molecules retain nearly all functional group types present in the ground-truth structures. 

To further verify the quality of the elucidated 3D molecular structures, we performed atom mapping against the corresponding ground-truth 3D structures using the Hungarian algorithm and computed the root mean square deviation (RMSD) in \AA. As shown in \cref{fig:performance}(b), when using a pre-trained SpecFormer with multi-modal spectra as input, DiffSpectra achieves the best elucidation performance, reaching 98.25\% atom mapping accuracy and an RMSD of 1.651. These results highlight the ability of our model to elucidate high-quality 3D structures with accurate atom types and precise coordinates.

We examined the detailed distributions of similarity-based metrics on the test set, as shown in \cref{fig:performance}(d). These metrics provide quantitative measures of how closely the elucidated structures resemble their corresponding ground-truth molecules. Across all metrics, \themodel, when equipped with a pre-trained SpecFormer and utilizing all spectral modalities as input, exhibits the most favorable distributions. This trend aligns well with the overall performance trends reported in \cref{fig:performance}(a).

We further analyzed the elucidation performance with respect to functional groups, as illustrated in \cref{fig:performance}(e–f). \cref{fig:performance}(e) presents the distribution of functional groups in the test molecules, showing a pronounced imbalance in their occurrence frequency. To address this imbalance, we evaluated model performance using the ROC-AUC score for each group, which reflects whether \themodel correctly predicts the presence or absence of a given functional group in each molecule. As shown in \cref{fig:performance}(f), \themodel achieves consistently high ROC-AUC values across diverse groups, demonstrating its ability to maintain reliable elucidation performance even under imbalanced data distributions.

Overall, these metrics collectively indicate that \themodel can accurately elucidate molecular structures from spectra, recovering not only global connectivity but also local functional groups and 3D geometry with high fidelity.

\begin{table*}[t]
\centering
\caption{
Structure elucidation performance of \themodel on the QM9S dataset. We compared two factors: (i) the effect of using a pre-trained versus an untrained SpecFormer as the spectral condition encoder, and (ii) the effect of generating molecular structures in 2D topology space versus in joint 2D\&3D geometry space. Reported metrics include exact Acc@$K$, MCES, fingerprint- and fragment-based similarities, functional group similarity, and, where applicable, 3D-specific measures such as RMSD (in \AA) and atom mapping accuracy.
}
\label{tab:specformer}
\scriptsize
\makebox[\textwidth][c]{
\resizebox{1.0\textwidth}{!}{
\begin{tabular}{cc|ccccccccc}
\toprule
\makecell{\textit{\textbf{Molecular}}\\\textit{\textbf{Modality}}} &\makecell{\textit{\textbf{Pre-trained}}\\\textit{\textbf{SpecFormer}}} & Acc@1 $\uparrow$ & MCES $\downarrow$ & $\operatorname{TaniSim}_{\mathrm{MG}}$ $\uparrow$ & $\operatorname{CosSim}_{\mathrm{MG}}$ $\uparrow$ & $\operatorname{TaniSim}_{\mathrm{MA}}$ $\uparrow$ & FraggleSim $\uparrow$ & FGSim $\uparrow$ & RMSD (\AA) $\downarrow$ & MapAcc $\uparrow$ \\
\midrule
\multirow{2}{*}{2D} & \xmark & 7.95\% & 2.6581   & 0.5295 & 0.6478 & 0.8144 & 0.8999 & 0.8930 & - & - \\
& \cmark & 9.49\% & 2.4132   & 0.5609 & 0.6734 & 0.8350 & 0.9072 & 0.9034 & - & - \\
\midrule
\multirow{2}{*}{2D\&3D} & \xmark & 35.64\% & 1.7795 & 0.7242 & 0.7964 & 0.8941 & 0.9397 & 0.9495 & 1.653 & 98.08\% \\
& \cmark & \first{40.76\%} & \first{1.3337} & \first{0.7875} & \first{0.8450} & \first{0.9262} & \first{0.9504} & \first{0.9652} & \first{1.651} & \first{98.25\%} \\
\bottomrule
\end{tabular}
}
}
\end{table*}

\subsection{Pre-trained SpecFormer facilitates more accurate structure elucidation}

To assess the impact of pre-training the spectrum encoder, we conducted an ablation study comparing DiffSpectra with and without a pre-trained SpecFormer module, as shown in \cref{fig:performance}, \cref{fig:visualization} and \cref{tab:specformer}. When equipped with the pre-trained SpecFormer, DiffSpectra achieves the top-1 accuracy of 40.76\%, compared to 35.64\% without pre-training, indicating a clear improvement in correctly recovering the ground-truth molecular structures from spectra. Similarly, the MCES decreases from 1.7795 to 1.3337, suggesting that molecular graphs predicted with pre-trained SpecFormer retain more of the correct connectivity of the target structures.

In addition, the fingerprint-based similarity scores also improve with pre-training. For example, the Tanimoto similarity over Morgan fingerprints rises from 0.7242 to 0.7875, and the cosine similarity improves from 0.7964 to 0.8450. The Tanimoto similarity over MACCS keys also increases from 0.8941 to 0.9262. These gains indicate that the molecular structures generated with pre-trained SpecFormer more accurately capture both local substructural patterns and functional-group motifs. Moreover, the Fraggle similarity improves from 0.9397 to 0.9504, and the FGSim from 0.9495 to 0.9652, confirming that larger chemical fragments and functional groups are better preserved.

Overall, these results demonstrate that pre-training the SpecFormer encoder on spectral data provides beneficial inductive biases, allowing DiffSpectra to more effectively align spectral representations with molecular graph structures during conditional generation. This highlights the value of leveraging a well-trained spectrum encoder to facilitate accurate molecular structure elucidation from spectra.

\subsection{Multi-modal spectra outperform single-modality spectra for structure elucidation}

To further investigate the contribution of different spectral modalities, we conducted an ablation study where \themodel was conditioned on individual or combined spectral types. For a fair comparison, the pre-trained SpecFormer was not used in these experiments. As shown in \cref{fig:performance} and \cref{fig:visualization}, using all spectral modalities (IR, Raman, UV-Vis) together achieves the best structure elucidation results, with a top-1 accuracy of 35.6\% and the lowest MCES of 1.780, indicating improved recovery of graph connectivity. The fingerprint-based metrics also show higher similarities when leveraging all spectra, with TaniSim$_\text{MG}$ at 0.7242 and CosSim$_\text{MG}$ at 0.7964, suggesting better reconstruction of local molecular patterns. Likewise, the Tanimoto similarity on MACCS keys reaches 0.8941, and the Fraggle similarity is 0.9397, indicating strong preservation of meaningful chemical fragments.

Among individual modalities, Raman spectra alone outperform IR and UV-Vis, with the top-1 accuracy of 30.3\% and reasonable similarity scores. IR spectra alone still provide useful chemical information, achieving 25.9\% top-1 accuracy, while UV-Vis alone performs poorly (0.3\% top-1 accuracy), reflecting its limited ability to uniquely identify molecular structures in the QM9S dataset. 

Overall, these results highlight that combining multiple spectral modalities provides complementary structural priors, enabling DiffSpectra to generate molecular structures with higher fidelity and greater consistency with ground-truth targets.

\subsection{Sampling multiple candidates improves structural hit accuracy}

Owing to the inherently stochastic nature of diffusion models, each sampling process may yield a distinct yet plausible molecular structure, even under the same conditions. Although a single sample may not always match the ground-truth structure exactly, generating multiple candidates increases the likelihood that at least one of them will align with the correct structure. This motivates the use of top-$K$ Accuracy (Acc@$K$) as a more comprehensive evaluation metric, formally defined in \cref{sec:metric2}.

We report in \cref{fig:performance}(C) the performance of different model variants across varying sample sizes. Specifically, Acc@$K$ quantifies the fraction of test molecules for which the correct structure appears within the top $K$ generated candidates.
For instance, the top-1 accuracy corresponds to the exact-match rate under single-sample generation, whereas Acc@$K$ ($K>1$) reflects the probability of recovering the correct structure when multiple guesses are permitted.
Across all model variants, 
Acc@$K$ exhibits a consistent monotonic increase with $K$, as the likelihood of including the correct structure naturally improves when more outputs are considered. Notably, the full model, when equipped with all spectra and pre-trained SpecFormer initialization, improves from 40.76\% at $K=1$ to 99.49\% at $K=10$. This demonstrates that even modest sampling budgets can yield near-exhaustive coverage of the true structure space.

These findings underscore an important property of diffusion-based molecular elucidation: although the model may not always recover the exact structure in a single attempt, it reliably ranks the correct structure within a compact set of plausible candidates. This property is particularly advantageous in practical workflows, where a ranked shortlist of candidates can be provided for subsequent experimental validation.

\subsection{Generating 3D geometries helps achieve more precise elucidation}

To assess the advantage of generating molecules in 3D geometry space, we compared structure elucidation in a purely 2D modality, where molecules are generated as topological graphs without spatial coordinates, against joint 2D and 3D generation. The results are summarized in \cref{tab:specformer}.

Incorporating 3D geometries yields substantial gains. This improvement arises because molecular spectra are intrinsically more correlated with 3D structures than with 2D topology: spectra reflect energy absorption patterns, whereas 3D structures more faithfully capture the energetic and spatial states of molecules. Consequently, the joint 2D and 3D generation achieves markedly higher exact elucidation accuracy (the top-1 accuracy increases from 9.49\% to 40.76\%), reduced MCES, and consistent improvements across fingerprint, fragment, and functional group similarities. By contrast, RMSD and mapping accuracy cannot be computed under the 2D-only settings, as 3D coordinates are absent.

Moreover, even within the 2D modality, the use of a pre-trained SpecFormer as the spectral condition encoder leads to measurable gains (e.g., the top-1 accuracy improves from 7.95\% to 9.49\%), highlighting the benefit of spectral pre-training.

\begin{table*}[h]
\centering
\caption{Evaluation of different SE(3) equivariance strategies for \themodel.
We compared a model-based equivariant architecture to data-based equivariance approaches, with or without data augmentation, on molecular structure elucidation metrics. Here, ``data aug'' refers to data augmentation.
}
\label{tab:equivariance}
\scriptsize
\makebox[\textwidth][c]{
\resizebox{1.00\textwidth}{!}{
\begin{tabular}{cccccccc}
\toprule
\textit{\textbf{SE(3) Equivariance}} & Acc@1 $\uparrow$ & MCES $\downarrow$ & $\operatorname{TaniSim}_{\mathrm{MG}}$ $\uparrow$ & $\operatorname{CosSim}_{\mathrm{MG}}$ $\uparrow$ & $\operatorname{TaniSim}_{\mathrm{MA}}$ $\uparrow$ & FraggleSim $\uparrow$ & FGSim $\uparrow$ \\
\midrule
model-based & \first{35.64\%} & \first{1.7795} & \first{0.7242} & \first{0.7964} & \first{0.8941} & \first{0.9397} & \first{0.9495} \\
data-based (w/ data aug) & 31.24\% & 1.9571 & 0.7025 & 0.7803 & 0.8834 & 0.9359 & 0.9420 \\
data-based (w/o data aug) & 20.25\% & 3.6036 & 0.5138 & 0.6308 & 0.7588 & 0.8658 & 0.8620 \\
\bottomrule
\end{tabular}
}
}
\end{table*}

\subsection{Comparing model-based and data-based SE(3) equivariance strategies}

Since our model performs diffusion generation in molecular 3D space, it is essential to ensure SE(3) equivariance so that predictions remain consistent under rigid-body transformations of the input. To this end, we designed and evaluated two strategies for achieving SE(3) equivariance: model-based equivariance and data-based equivariance.
In the model-based approach, geometric inductive biases are explicitly incorporated into the network architecture, for example, through pairwise distance encodings and equivariant coordinate updates, which ensure equivariant behavior. By contrast, the data-based approach does not include SE(3)-equivariant inductive biases in the model architecture. Instead, it relies on random rotation and translation augmentations during training, encouraging the model to learn equivariance empirically from data.

As shown in \cref{tab:equivariance}, the model-based SE(3)-equivariant implementation of \themodel achieves consistently superior performance. It attains the highest top-1 accuracy (35.64\%) and the lowest MCES (1.7795), indicating stronger recovery of molecular graph connectivity. The data-based approach with data augmentation achieves slightly lower top-1 accuracy (31.24\%). Notably, when data augmentation is removed from the data-based approach, performance degrades substantially across all metrics, confirming the critical role of data augmentation in enabling the model to generalize over molecular rotations and translations. 
Overall, these results demonstrate that while data-based equivariance with augmentation can help achieve some degree of SE(3)-aware behavior, explicitly incorporating geometric inductive biases through model-based equivariance constitutes a more robust and effective strategy for molecular structure elucidation from spectra.

\subsection{Generalization across molecules with different numbers of atoms}

\begin{wrapfigure}[18]{r}{0.43\linewidth}
  \centering
  \vspace{-1.3em}
  \includegraphics[width=\linewidth]{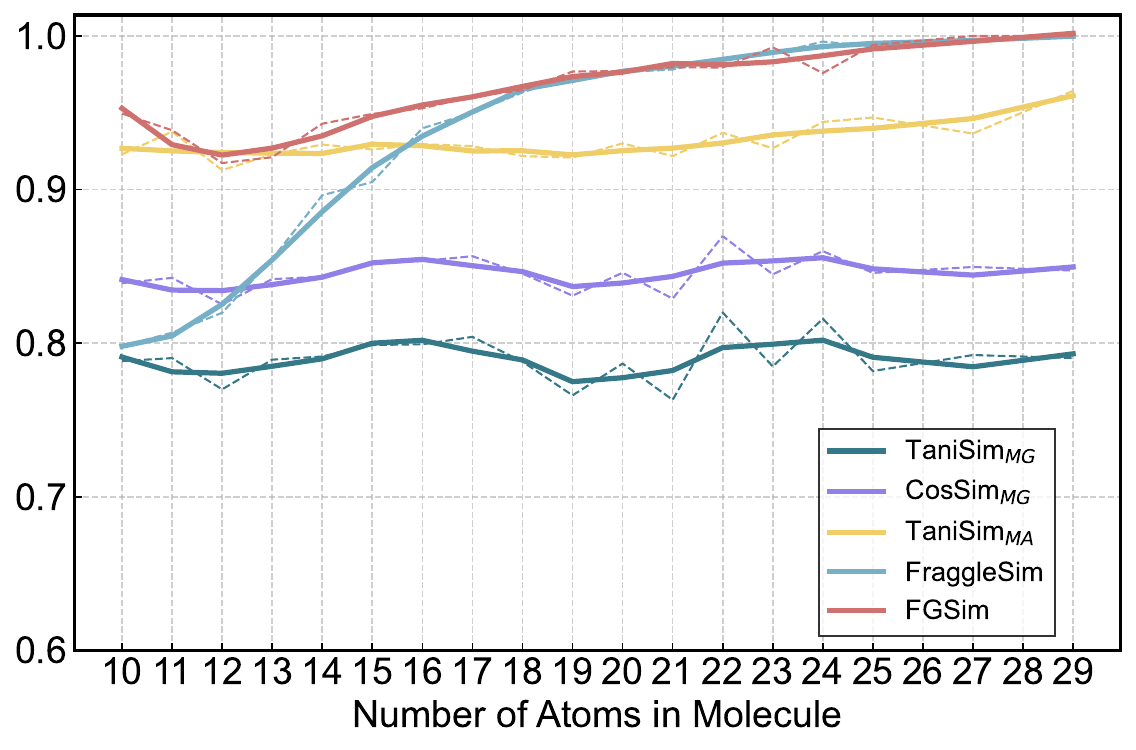}
  \vspace{-1em}
  \caption{Molecular structure elucidation performance across molecules with varying numbers of atoms. The figure presents results obtained using different similarity metrics, including $\operatorname{TaniSim}_{\mathrm{MG}}$, $\operatorname{CosSim}_{\mathrm{MG}}$, $\operatorname{TaniSim}_{\mathrm{MA}}$, FraggleSim, and FGSim.}
  \label{fig:atom_nums}
\end{wrapfigure}

To evaluate the generalizability of \themodel across molecules of varying sizes, we analyzed its performance on molecules containing different numbers of atoms. \cref{fig:atom_nums} presents the results across multiple similarity metrics, where dashed lines denote raw similarity values and solid curves represent the smoothed trends obtained using the Savitzky-Golay filter~\cite{SG-Filter}.

As shown in \cref{fig:atom_nums}, the similarity metrics based on molecular fingerprints, such as $\mathrm{TaniSim}_{\mathrm{MG}}$, $\mathrm{CosSim}_{\mathrm{MG}}$, and $\mathrm{TaniSim}_{\mathrm{MA}}$, remain consistently stable across molecules of different sizes. This indicates that DiffSpectra achieves robust generalization regardless of molecular size. In contrast, fragment- and functional group-based similarity metrics (FraggleSim and FGSim) exhibit slightly lower performance when the number of atoms is small. However, as the number of atoms increases, both metrics steadily improve and approach values close to 1.0. This trend arises because very small molecules cannot be effectively decomposed into representative fragments or functional groups, which constrains the reliability of these similarity measures. With larger molecular structures, the decomposition becomes more informative, resulting in improved and stable performance.

In summary, these results demonstrate that DiffSpectra maintains reliable generalization across molecular scales, ensuring consistent performance for molecules of different sizes.

\subsection{Elucidation trajectory analysis based on gradient}

\begin{figure}
\begin{center}
\includegraphics[width=\linewidth]{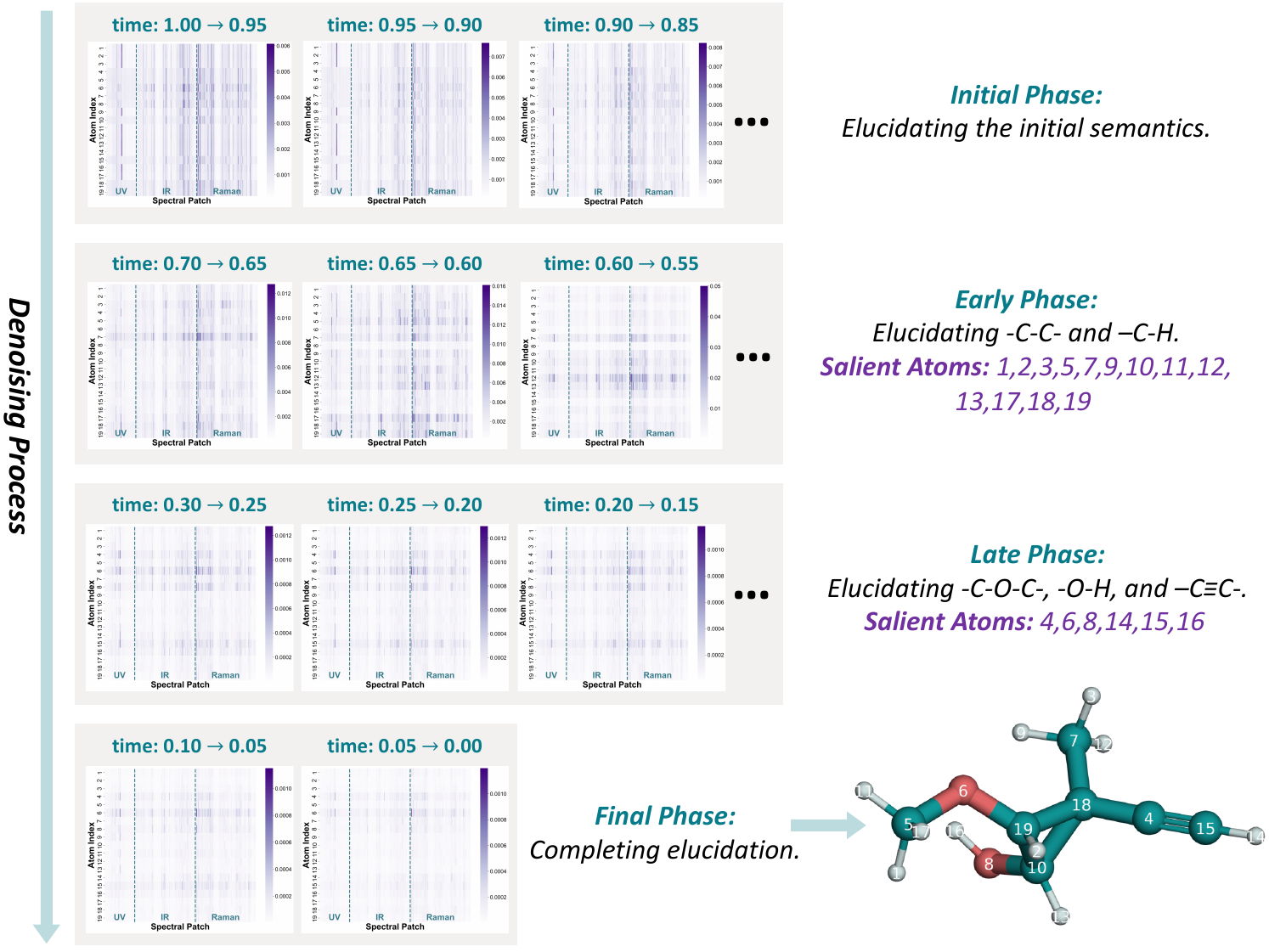}
\end{center}
\caption{\textbf{Gradient-based elucidation trajectory analysis.}
Gradient heatmaps depict atomic gradients with respect to spectral patches, with each heatmap averaged over a time interval of 0.05 (corresponding to 50 denoising steps). 
Upon completion of the denoising process, a molecular structure is elucidated that matches the ground-truth structure.
Atom indices are annotated on the structure.
}
\label{fig:gradient}
\end{figure}

To further reveal the generative trajectory of \themodel and gain deeper insight into its structural elucidation process, we analyzed the gradients of each atom with respect to the input spectral patches. Specifically, we computed the gradient of the predicted atom type for each atom with respect to the input patches at each denoising step. Since the denoising process in \themodel consists of 1,000 steps, this procedure yields a total of 1,000 gradient heatmaps. In these heatmaps, each row corresponds to an indexed atom, while each column corresponds to a spectral patch. Larger gradient values indicate that, at the corresponding time step, a spectral patch exerts a stronger influence on the prediction of the atom's type. To better visualize the generative trajectory, we averaged the gradients over every 50 steps, corresponding to time intervals of 0.05 along the generative time axis (ranging from 1 to 0). The resulting averaged gradient heatmaps, together with the final molecular structure annotated by atom indices, are presented in \cref{fig:gradient}.

Inspection of these heatmaps reveals that the generative behavior of DiffSpectra evolves dynamically throughout the denoising process. Based on these observations, we delineate the trajectory into four characteristic phases: initial, early, late, and final phases. During the initial phase (approximately from 1.0 to 0.7), the gradients of all atoms are elevated, suggesting that the model is establishing a broad semantic representation that provides the foundation for subsequent, more fine-grained assignments. During the early phase (approximately from 0.7 to 0.3), atoms within simple substructures, particularly \ce{-C-C-} and \ce{-C-H} bonds, exhibit stronger gradients, indicating that these elementary motifs are elucidated first. The late phase (approximately from 0.3 to 0.1) is marked by the preferential elucidation of more complex functional groups, such as ethers (\ce{-C-O-C-}), hydroxyls (\ce{-OH}), and alkynes (\ce{-C#C-}), which are assigned higher gradient values during this phase. Finally, in the final phase (approximately from 0.1 to 0.0), the gradients associated with all atoms diminish gradually, signifying that \themodel has largely completed the elucidation process and converged on a confident molecular structure.

In summary, gradient-based trajectory analysis reveals that \themodel first generates initial semantics encompassing all atoms, then elucidates simple substructures, subsequently focuses on more complex functional groups, and finally consolidates the molecular structure with high confidence.

\begin{table*}[h]
\centering
\caption{Effect of sampling temperature on structure elucidation performance.
We report \themodel results with different sampling temperature coefficients $\tau$, which scale the injected noise during diffusion sampling. 
Moderate values of $\tau$ help balance diversity and accuracy, while extremely low or high $\tau$ degrade performance.}
\label{tab:temperature}
\scriptsize
\makebox[\textwidth][c]{
\resizebox{1\textwidth}{!}{
\begin{tabular}{ccccccccc}
\toprule
\textit{\textbf{Temperature}} & Acc@1 $\uparrow$ & MCES $\downarrow$ & $\operatorname{TaniSim}_{\mathrm{MG}}$ $\uparrow$ & $\operatorname{CosSim}_{\mathrm{MG}}$ $\uparrow$ & $\operatorname{TaniSim}_{\mathrm{MA}}$ $\uparrow$ & FraggleSim $\uparrow$ & FGSim $\uparrow$ \\
\midrule
$\tau$=0.0 & 14.65\% & 4.6948 & 0.4408 & 0.5595 & 0.6754 & 0.8136 & 0.7819 \\
$\tau$=0.2 & 33.23\% & 1.9900 & 0.7005 & 0.7774 & 0.8763 & 0.9297 & 0.9354 \\
$\tau$=0.4 & 38.31\% & 1.4846 & 0.7659 & 0.8287 & 0.9151 & 0.9449 & 0.9595 \\
$\tau$=0.6 & 40.52\% & 1.3694 & 0.7826 & 0.8414 & 0.9232 & 0.9500 & 0.9637 \\
$\tau$=0.8 & 40.40\% & \first{1.3190} & 0.7859 & 0.8438 & 0.9250 & \first{0.9512} & 0.9651 \\
$\tau$=1.0 & \first{40.76\%} & 1.3337 & \first{0.7875} & \first{0.8450} & \first{0.9262} & 0.9504 & \first{0.9652} \\
$\tau$=1.2 & 39.59\% & 1.3850 & 0.7811 & 0.8402 & 0.9224 & 0.9485 & 0.9628 \\
\bottomrule
\end{tabular}
}
}
\end{table*}

\subsection{Sampling temperature balances diversity and accuracy}

Since each molecular spectrum uniquely determines a molecular structure, accurate molecular structure elucidation is paramount. However, diffusion models inherently introduce stochasticity during sampling through the injection of random noise, which is essential for promoting diversity in the generated results. To better understand the trade-off between stochasticity and accuracy, we examined how modulating this randomness affects performance. Specifically, we introduce a sampling temperature parameter $\tau$ in \cref{eq:sampling}, which scales the injected noise during sampling. Lower values of $\tau$ diminish stochasticity, resulting in more deterministic outputs, while higher values of $\tau$ allow for broader exploration of the molecular space.

As shown in \cref{tab:temperature}, we evaluated DiffSpectra under varying $\tau$ values ranging from 0.0 to 1.2. Moderate temperature values ($\tau=0.8$ and $\tau=1.0$) achieve the best balance between diversity and accuracy. In particular, $\tau=1.0$ obtains the highest top-1 accuracy of 40.76\%, while maintaining competitive similarity metrics, including a Tanimoto$_{\mathrm{MG}}$ of 0.7875, Cosine$_{\mathrm{MG}}$ of 0.8450, and Tanimoto$_{\mathrm{MA}}$ of 0.9262. Extremely low temperatures (e.g., $\tau=0.0$) result in overly deterministic results with reduced accuracy and elevated MCES, while extremely high temperatures (e.g., $\tau=1.2$) also degrade performance due to excessive randomness.

Taken together, these results suggest that a moderate choice of $\tau$ helps DiffSpectra strike a balance between structural diversity and high fidelity to the ground-truth molecules.
\section{Discussion}\label{sec:discussion}

In this work, we introduced DiffSpectra, the first spectra-conditioned diffusion framework for molecular structure elucidation that directly integrates multi-modal spectral information into both 2D topological and 3D geometric structure generation. 
Our results demonstrate that the framework not only ensures chemical validity and stability that are comparable to the training distribution but also achieves a top-1 exact accuracy of 40.8\%, with functional group recovery exceeding 96\%. Furthermore, by sampling multiple candidates, \themodel consistently ranks the correct structure among the top 10 predictions, yielding a top-10 accuracy of 99.5\%. This highlights the practical utility of diffusion-based sampling, whereby a shortlist of candidates can be reliably provided for downstream validation.

Several core findings underpin the effectiveness of DiffSpectra. First, the incorporation of 3D geometries alongside topological graphs markedly improves structure elucidation performance, underscoring the fact that spectral signatures are inherently tied to molecular geometry. Second, our comparison of SE(3) equivariance strategies shows that explicitly embedding geometric inductive biases into the model architecture provides more robust performance than relying solely on data augmentation, emphasizing the importance of principled symmetry constraints. Third, conditioning on multiple spectral modalities leads to complementary structural information: while Raman spectra provide richer constraints than UV-Vis, the integration of Raman, IR, and UV-Vis yields the most accurate elucidations. Fourth, pre-training the spectral encoder SpecFormer with masked reconstruction and contrastive alignment proves highly beneficial, providing inductive biases that significantly improve spectral–structural alignment. Finally, gradient-based trajectory analysis reveals a mechanistic view of the generative process: DiffSpectra first establishes global semantic information across atoms, then elucidates simple substructures, subsequently resolves more complex functional groups, and ultimately converges to a chemically consistent molecular structure. This staged generative trajectory mirrors the reasoning process of human chemists and provides interpretability for the model’s predictions.

The demonstrated performance and interpretability of DiffSpectra have direct implications for practical applications in natural product discovery, drug development, and functional materials design. By providing high-quality candidate structures directly from spectra, DiffSpectra offers a scalable alternative to traditional manual elucidation pipelines and opens opportunities for integrating automated generative models into real-world experimental workflows.

Despite its encouraging performance, DiffSpectra currently focuses on three spectral types—UV-Vis, IR, and Raman—so its applicability to other modalities, such as nuclear magnetic resonance (NMR) or mass spectrometry (MS), remains to be further explored. In addition, our validation to date has been primarily computational. Therefore, extensive experimental studies will be an important next step toward further strengthening the robustness of the framework in practical laboratory settings.

Our work paves the way for several promising research directions. Expanding the diversity and scale of available spectral datasets will likely further improve the generalizability and performance of spectra-conditioned generation. Extending the framework to handle more complex modalities such as NMR, MS, or X-ray spectra would enhance its applicability across a wider range of analytical chemistry tasks. Moreover, adapting DiffSpectra to larger biomolecular and materials systems could unlock new opportunities for molecular identification in drug discovery and materials science, particularly when coupled with high-throughput experimental pipelines. 

In summary, DiffSpectra represents a significant advance in molecular structure elucidation by unifying topological, geometric, and spectral reasoning within a diffusion-based generative framework. By demonstrating high accuracy, interpretability, and generalizability, DiffSpectra establishes a new paradigm for automated and scalable structure elucidation. We believe that this work makes an important step toward trustworthy and accurate spectra-conditioned molecular structure elucidation, with broad potential to impact computational chemistry, drug discovery, and materials science.
\section{Methods}\label{sec:method}

\subsection{Diffusion Model for Molecular Structure Elucidation}

We adopt a continuous-time diffusion probabilistic model to generate molecular graphs in a joint space that integrates 2D topology and 3D geometry. Each molecule is represented as a graph $\mathcal{G} = (\mathbf{H}, \mathbf{A}, \mathbf{X})$, where $\mathbf{H} \in \mathbb{R}^{N \times d_1}$ denotes node-level attributes such as atom types and charges, $\mathbf{A} \in \mathbb{R}^{N \times N \times d_2}$ encodes pairwise edge features such as bond existence and bond types, and $\mathbf{X} \in \mathbb{R}^{N \times 3}$ corresponds to the 3D coordinates of the atoms. Here, $N$ represents the number of atoms in the molecule, $d_1$ denotes the dimensionality of the node features, and $d_2$ denotes the dimensionality of the edge features. 

We adopt the mathematical formulation of the Variational Diffusion Model (VDM)~\citep{VDM,ProgressiveDistillation,BlurringDiffusion} to define our diffusion and sampling processes, which follow a continuous-time diffusion probabilistic framework, as illustrated in \cref{fig:overview}(A). To streamline the presentation, we consider a generic graph component, namely the node features $\mathbf{H}$, the edge features $\mathbf{A}$, or the atomic coordinates $\mathbf{X}$, which we uniformly denote by a vector-valued variable $\mathbf{G} \in \mathbb{R}^d$. The subsequent formulas can then be applied independently to each component in a consistent manner.

\subsubsection{Forward Diffusion Process}

In the forward diffusion process, Gaussian noise $\bG_\epsilon \sim \mathcal{N}(\b0, \bI)$ is gradually added to the data over continuous time from $t=0$ to $t=1$. The noised sample $\bG_t$ is obtained by:
\begin{equation}
    \bG_t = \alpha_t \bG_0 + \sigma_t \bG_\epsilon,
\end{equation}
where $\alpha_t$ and $\sigma_t$ are signal and noise scaling functions, typically determined by cosine or linear schedules. The signal-to-noise ratio (SNR) at time $t$ is defined as:
\begin{equation}
    \mathrm{SNR}(t) = \frac{\alpha_t^2}{\sigma_t^2}.
\end{equation}
The SNR is strictly decreasing from $t = 0$ to $t = 1$, ensuring that the data is gradually corrupted into pure noise as $t \to 1$.
Given two time steps, $s$ and $t$, $0 \leq s < t \leq 1$, the conditional distribution of a noised graph at time $t$ given a less-noised graph at time $s$, denoted $q(\bG_t \mid \bG_s)$, is also Gaussian:
\begin{equation}
    q(\bG_t \mid \bG_s) = \mathcal{N}(\alpha_{t|s} \bG_s, \sigma_{t|s}^2 \bI),
\end{equation}
where the intermediate scaling factor and conditional variance are defined as:
\begin{equation}
    \alpha_{t|s} = \frac{\alpha_t}{\alpha_s}, \quad
    \sigma_{t|s}^2 = \sigma_t^2 - \alpha_{t|s}^2 \sigma_s^2.
\end{equation}
This formulation allows the forward process to be implemented using analytically tractable Gaussian transitions, which also facilitates efficient computation of training objectives and reverse-time sampling.

\subsubsection{Reverse Denoising Process}
The reverse-time denoising process starts from a standard Gaussian sample $\bG_1 \sim \mathcal{N}(\b0, \bI)$ and proceeds backward from $t = 1$ to $t = 0$. Let $0 \leq s < t \leq 1$ denote two consecutive time steps. At each step, a data prediction model $d_\theta$ takes as input the noised sample $\bG_t$, a self-conditioning~\cite{SelfConditioning} estimate $\tilde{\bG}_0$, and the log signal-to-noise ratio $\log \mathrm{SNR}(t)$. The model predicts $\hat{\bG}_0 = d_\theta(\bG_t, \tilde{\bG}_0, \log \mathrm{SNR}(t))$, which is then used to compute the denoised sample $\bG_s$ by:
\begin{gather}
    \bar{\bG}_s = \frac{\alpha_{t|s} \sigma_s^2}{\sigma_t^2} \bG_t + \frac{\alpha_s \sigma_{t|s}^2}{\sigma_t^2} \hat{\bG}_0, \\
    \bG_s = \bar{\bG}_s + \tau \cdot \frac{\sigma_s \sigma_{t|s}}{\sigma_t} \bG_\epsilon, \quad \bG_\epsilon \sim \mathcal{N}(\b0, \bI),
    \label{eq:sampling}
\end{gather}
where $\tau$ is a sampling temperature parameter, modulating the amount of stochasticity in the sampling process. Lower values of $\tau < 1$ reduce the influence of noise, leading to more deterministic outputs.
The self-conditioning mechanism, where the model reuses its previous prediction $\tilde{\bG}_0$ as an additional input in the next step, enhances training stability and generation quality~\cite{SelfConditioning}.

\subsubsection{Training Objective}
To recover the original sample, we train a data prediction model $d_\theta$ that takes as input the noised graph $\mathbf{G}_t$, a self-conditioning estimate $\tilde{\mathbf{G}}_0$, and the log signal-to-noise ratio $\log \mathrm{SNR}(t)$. The model predicts the clean sample $\hat{\mathbf{G}}_0$, where $\hat{\mathbf{G}}_0$ refers to any component of $\hat{\mathcal{G}}_0 = (\hat{\mathbf{H}}, \hat{\mathbf{A}}, \hat{\mathbf{X}})$. Therefore, the model is trained to minimize a weighted mean squared error:
\begin{equation}
    \mathcal{L} = \mathbb{E}_{t, \mathbf{G}_0} \left[ \sqrt{\frac{\alpha_t}{\sigma_t}} \left( \lambda_1 \| \hat{\mathbf{A}} - \mathbf{A}_0 \|_2^2 + \lambda_2 \| \hat{\mathbf{X}} - \hat{\mathbf{X}}_0 \|_2^2 + \lambda_3 \| \hat{\mathbf{H}} - \mathbf{H}_0 \|_2^2 \right) \right],
\end{equation}
where $\hat{\mathbf{X}}_0$ is obtained by aligning $\mathbf{X}_0$ to $\mathbf{X}_t$ using the Kabsch algorithm~\cite{KabschAlign} to preserve SE(3)-equivariance. The loss coefficients $\lambda_i$ balance the relative importance of each component.

\subsection{Diffusion Molecule Transformer (DMT)}
To parameterize the data prediction module $d_\theta$, we employ a specialized architecture, the \emph{Diffusion Molecule Transformer} (DMT)~\cite{JODO,NExT-Mol}, as its backbone, to jointly model the node features, edge features, and 3D coordinates of  the noisy molecular graph within our diffusion framework. The architecture is motivated by the need to recover the correlations among these three components, each of which is independently corrupted by noise during the forward diffusion process.
DMT is specifically tailored for molecular generation tasks and preserves both permutation equivariance and SE(3)-equivariance.

At each time $t \in [0,1]$ during the reverse process, DMT takes the noisy molecular graph $\mathcal{G}_t = (\mathbf{H}_t, \mathbf{A}_t, \mathbf{X}_t)$ as input, along with a self-conditioned estimate $\tilde{\mathcal{G}}_0$, and predicts a clean graph $\hat{\mathcal{G}}_0 = (\hat{\mathbf{H}}, \hat{\mathbf{A}}, \hat{\mathbf{X}})$. The model performs denoising inference conditioned on both the current time step and the molecular spectra. The overall model architecture is illustrated in \cref{fig:overview}(B). For the sake of clarity, the self-conditioning mechanism is omitted in the figure.

A time embedding, obtained via sinusoidal encoding of $\log(\alpha_t^2/\sigma_t^2)$, is passed through learnable projections. Simultaneously, embeddings of spectra are generated by the SpecFormer, which will be detailed in \cref{sec:specformer}. These two embeddings are then concatenated to form the conditioning vector $\bC$, which is used throughout the network.

\textbf{Initial embeddings.}
The DMT module consists of stacked Transformer-style equivariant blocks. In the first layer, the node features $\bH_t$ and the edge features $\bA_t$ are concatenated with their corresponding self-conditioning estimates $\tilde{\bH}_0$ and $\tilde{\bA}_0$, respectively. For edge features, distance features $\mathbf{D}_0$ computed from the self-conditioned coordinates $\tilde{\bX}_0$ are also concatenated. These concatenated features are then passed through linear projections to obtain the initial embeddings:
\begin{equation}
\bH^{(1)} = \mathrm{Linear}_{\mathrm{node}}\bigl(\left[\bH_t, \tilde{\bH}_0\right]\bigr) \in \mathbb{R}^{N \times d_{h}},
\end{equation}
\begin{equation}
\bA^{(1)} = \mathrm{Linear}_{\mathrm{edge}}\bigl(\left[\bA_t, \tilde{\bA}_0, \mathbf{D}_0\right]\bigr) \in \mathbb{R}^{N \times N \times d_{a}},
\end{equation}
where $\left[\cdot,\cdot\right]$ denotes the concatenation operation, and $d_{h}$ and $d_{a}$ denote the hidden embedding dimensions for node and edge embeddings, respectively. 
The coordinate stream initializes with the noisy positions $\bX^{(1)} = \bX_t$. In addition, adjacency matrices derived from chemical bonding patterns and distance cutoffs computed from $\tilde{\bX}_0$ are extracted and concatenated to serve as auxiliary attention masks in subsequent layers.

\textbf{Three-stream update.}
Each DMT block consists of three interacting streams responsible for updating node features, edge features, and coordinates, respectively. These streams exchange information throughout the block. In the node stream, information is propagated across the molecular graph via a relational multi-head attention (MHA) mechanism~\cite{GTSurvey,GSLB}, enabling flexible message passing over fully connected molecular graphs. In particular, pairwise geometric distances computed from coordinates are transformed into radial basis encodings $\rho_{ij}^{(l)}$, and these are combined with edge features and geometric distances such as:
\begin{equation}
\bar{\bA}_{ij}^{(l)} = \left[ \bA_{ij}^{(l)};\, \|\bX^{(l)}_i - \bX^{(l)}_j\|_2;\, \rho_{ij}^{(l)} \right],
\end{equation}
forming a geometry-aware relational representation. The queries, keys, and values for MHA are obtained by applying learnable linear projections to node features:
\begin{equation}
\bQ = \bH^{(l)} \bW^Q, \quad
\bK = \bH^{(l)} \bW^K, \quad
\bV = \bH^{(l)} \bW^V,
\end{equation}
where $\bW^Q, \bW^K, \bW^V$ are learned projection matrices. The attention weights are then defined as:
\begin{equation}
a_{ij} = \frac{ \tanh\bigl(\phi_0(\bar{\bA}_{ij}^{(l)})\bigr)\, \bQ_i\, \bK_j^\top }{ \sqrt{d_k} },
\quad
a = \mathrm{Softmax}(a),
\end{equation}
and the node-level aggregation proceeds as:
\begin{equation}
\mathrm{Attention}(\bH^{(l)}, \bar{\bA}^{(l)})_i
= \sum_{j=1}^N a_{ij}\, \tanh\bigl(\phi_1(\bar{\bA}_{ij}^{(l)})\bigr)\, \bV_j,
\end{equation}
with $\phi_0(\cdot)$ and $\phi_1(\cdot)$ as learnable projections. 
We extend the above attention mechanism to a multi-head relational attention framework. 
For each head $h = 1, \dots, H$, we compute head-specific queries, keys, and values as:
\begin{equation}
\bQ_h = \bH^{(l)} \bW_h^Q,\quad 
\bK_h = \bH^{(l)} \bW_h^K,\quad
\bV_h = \bH^{(l)} \bW_h^V,
\end{equation}
where $\bW_h^Q, \bW_h^K, \bW_h^V$ are learnable projection matrices for head $h$. Each head independently performs geometry-aware attention as:
\begin{equation}
\bM_h^{(l)} = \mathrm{Attention}_h\bigl(\bH^{(l)}, \bar{\bA}^{(l)}\bigr),
\end{equation}
where $\mathrm{Attention}_h(\cdot)$ denotes the relational attention operator defined above. The outputs from all heads are then concatenated and projected back to the model dimension:
\begin{equation}
\bM^{(l)} = \mathrm{Concat}\bigl[\bM_1^{(l)}, \dots, \bM_H^{(l)}\bigr].
\end{equation}
The resulting $\bM^{(l)} \in \mathbb{R}^{N \times d_m}$ serves as the aggregated node embedding enriched by geometry-aware multi-head interactions.

To incorporate time-dependent and spectra-dependent conditioning, we apply adaptive layer normalization (AdaLN)~\cite{FiLM,DiT} to each stream, conditioned on a learned conditioning embedding $\mathbf{C}$: $\mathrm{AdaLN}(\vh, \mathbf{C}) = \left( 1 + \mathrm{FFN}_{\mathrm{scale}}(\mathbf{C}) \right) \cdot \mathrm{LN}(\vh) + \mathrm{FFN}_{\mathrm{bias}}(\mathbf{C})$,
where $\mathrm{FFN}_{\mathrm{scale}}(\cdot)$ and $\mathrm{FFN}_{\mathrm{bias}}(\cdot)$ are lightweight feed-forward networks that project the conditioning signal into scale and shift parameters.
In addition, we introduce an adaptive scaling function to further modulate feature magnitude: $\mathrm{Scale}(\vh, \mathbf{C}) = \mathrm{FFN}_{\mathrm{scale}}'(\mathbf{C}) \cdot \vh$, where $\mathrm{FFN}_{\mathrm{scale}}'(\cdot)$ shares the same architecture as the above FFNs but produces a multiplicative gating factor.
The node update is performed by:
\begin{equation}
\bH^{(l+1)'} = \mathrm{Scale}\bigl(\bM^{(l)}, \mathbf{C}\bigr) + \bH^{(l)},
\end{equation}
\begin{equation}
\bH^{(l+1)} = \mathrm{Scale}\left(
  \mathrm{FFN}\bigl( \mathrm{AdaLN}(\bH^{(l+1)'}, \mathbf{C})\bigr),
  \mathbf{C}
\right) + \bH^{(l+1)'}.
\end{equation}
Similarly, edge features are updated by first fusing node messages:
\begin{equation}
\hat{\bA}_{ij}^{(l)} = \left( \bM^{(l)}_i + \bM^{(l)}_j \right) \bW_1,
\end{equation}
with subsequent normalization and scaling:
\begin{equation}
\bA^{(l+1)'}_{ij} = \mathrm{Scale}\bigl( \hat{\bA}_{ij}^{(l)}, \mathbf{C} \bigr) + \bA^{(l)}_{ij},
\end{equation}
\begin{equation}
\bA^{(l+1)}_{ij} = \mathrm{Scale}\left(
  \mathrm{FFN}\bigl( \mathrm{AdaLN}(\bA^{(l+1)'}_{ij}, \mathbf{C})\bigr),
  \mathbf{C}
\right) + \bA^{(l+1)'}_{ij}.
\end{equation}
The parameter $\bW_1$ is a learnable weight matrix that couples node and edge information.  

For the coordinate stream, equivariant updates are achieved using directional vector fields, combining signals from node and edge streams:
\begin{equation}
e_{ij}^{(l+1)} = \mathrm{AdaLN}\left(
  \bW_2
  \left[
    \bH_i^{(l+1)},
    \bH_j^{(l+1)},
    \bA_{ij}^{(l+1)},
    \|\bX^{(l)}_i - \bX^{(l)}_j\|_2
  \right],
  \mathbf{C}
\right),
\end{equation}
\begin{equation}
\bX_i^{(l+1)} = \bX^{(l)}_i + \sum_{j \ne i} \gamma^{(l)} \,
\frac{ \bX^{(l)}_i - \bX^{(l)}_j }
     { \|\bX^{(l)}_i - \bX^{(l)}_j\|_2 }
\, \tanh\bigl(\mathrm{FFN}( e_{ij}^{(l+1)} )\bigr),
\end{equation}
where $\bW_2$ is a learned projection and $\gamma^{(l)}$ is a trainable scalar that stabilizes the directional updates. Finally, coordinates are shifted to have zero center-of-mass by mean centering, ensuring translation invariance.

Overall, the DMT stacks $L$ such equivariant blocks, which collectively refine the graph features and spatial coordinates. Final output heads predict discrete atom and bond types as well as coordinates aligned to the canonical centered frame. Its design guarantees permutation equivariance and SE(3) equivariance, enabling robust molecule generation across both topological and geometric levels.

\subsection{Spectra Transformer (SpecFormer) for Spectra Encoding}\label{sec:specformer}

To enable DMT to incorporate molecular spectral information as a conditional input for molecular structure elucidation, we propose the Spectra Transformer (SpecFormer)~\cite{MolSpectra} to encode molecular spectra. We further pre-train SpecFormer using both molecular structure data and spectra data.

\subsubsection{Architecture of SpecFormer}

For each type of spectrum, we first segment the data into patches and  independently encode these patches. The patch embeddings from all spectra are then concatenated and collectively processed by a Transformer-based encoder.

\textbf{Patching.}
Rather than encoding each frequency point individually, we divide each spectrum into multiple patches. 
This design choice offers two key benefits: (\romannumeral 1) it enables the model to capture local semantic features—such as absorption peaks—more effectively by grouping adjacent frequency points; and (\romannumeral 2) it reduces the computational cost for the following Transformer layers. 
Formally, a spectrum $\vs_i \in \mathbb{R}^{L_i}$ (where $i = 1, \dots, |\mathcal{S}|$) is split into patches of length $P_i$ with stride $D_i$. If $0 < D_i < P_i$, patches overlap by $P_i - D_i$ points; if $D_i = P_i$, patches are non-overlapping. The patching process yields a sequence $\vp_i \in \mathbb{R}^{N_i \times P_i}$, where $N_i = \left\lfloor \frac{L_i - P_i}{D_i} \right\rfloor + 1$ is the number of patches produced from $\vs_i$.

\textbf{Patch and position encoding.}
Each patch sequence from the $i$-th spectrum is projected into a latent space of dimension $d$ via a learnable linear transformation $\bW_i \in \mathbb{R}^{P_i \times d}$. To preserve the sequential order, a learnable positional encoding $\bW_i^{\text{pos}} \in \mathbb{R}^{N_i \times d}$ is added: $\vp_i' = \vp_i \bW_i + \bW_i^{\text{pos}}$, yielding the encoded representation $\vp_i' \in \mathbb{R}^{N_i \times d}$ for each spectrum. These representations serve as inputs to the SpecFormer encoder.

\textbf{SpecFormer: multi-spectrum Transformer encoder.}
Existing encoders, such as CNN-AM~\citep{CNN-AM}, utilize one-dimensional convolutions and are typically designed to process a single type of spectrum. In contrast, our model simultaneously considers multiple molecular spectra (e.g., UV-Vis, IR, Raman) as input. This multi-spectrum strategy is motivated by the presence of both \textit{intra-spectrum dependencies}—relationships among peaks within the same spectrum—and \textit{inter-spectrum dependencies}—correlations between peaks across different spectral modalities. These dependencies have been well documented, for example, in studies on vibronic coupling~\citep{Vibronic-Coupling}.

To effectively model these dependencies, we concatenate the encoded patch sequences from all spectra: $\hat{\vp} = \vp_1' \| \cdots \| \vp_{|\mathcal{S}|}' \in \mathbb{R}^{(\sum_{i=1}^{|\mathcal{S}|} N_i) \times d}$. This combined sequence is then passed through a Transformer encoder (see \cref{fig:overview}). In each attention head $h = 1, \ldots, H$, queries, keys, and values are computed as $\bQ_h = \hat{\vp} \bW_h^Q$, $\bK_h = \hat{\vp} \bW_h^K$, and $\bV_h = \hat{\vp} \bW_h^V$ respectively, with $\bW_h^Q, \bW_h^K \in \mathbb{R}^{d \times d_k}$ and $\bW_h^V \in \mathbb{R}^{d \times \frac{d}{H}}$. The attention output for each head is:
\begin{equation}
\bO_h = \text{Attention}(\bQ_h, \bK_h, \bV_h) = \text{Softmax}\left(\frac{\bQ_h \bK_h^{\top}}{\sqrt{d_k}}\right)\bV_h.
\end{equation}
Batch normalization, feed-forward networks, and residual connections are integrated as illustrated in \cref{fig:overview}. The outputs from all attention heads are combined to form $\vz \in \mathbb{R}^{(\sum_{i=1}^{|\mathcal{S}|} N_i) \times d}$. Finally, a flattening layer and a projection head are applied to produce the molecular spectra representation $\vz_s \in \mathbb{R}^d$.

\subsubsection{Pre-training of SpecFormer}

To enable more effective encoding of molecular spectra, we introduce a masked-reconstruction pre-training objective. In addition, to mitigate the scarcity of spectral pre-training data by leveraging large-scale molecular structure pre-training, we incorporate a contrastive learning objective to align spectral and structural representations. The complete pre-training framework is illustrated in \cref{fig:overview}(C).

\textbf{Masked patches reconstruction pre-training for spectra.}
To ensure that the spectrum encoder can effectively extract and represent information from molecular spectra, we adopt a masked patches reconstruction (MPR) pre-training strategy. Inspired by the effectiveness of masked reconstruction across multiple domains~\citep{Bert,MAE,GraphMAE,Mole-Bert,AUG-MAE,PatchTST}, MPR guides the learning of SpecFormer by encouraging it to reconstruct the masked portions of the spectral data.

After segmenting the spectra into patches, we randomly mask a fraction of these patches—according to a predefined ratio $\alpha$—by replacing them with zero vectors. The masked patch sequences then undergo patch and positional encoding, which conceals their semantic content (such as absorption intensities at certain wavelengths) but retains their positional information, thereby aiding the reconstruction task.

After processed by SpecFormer, the encoded representations corresponding to the masked patches are passed through a reconstruction head specific to each spectrum. The original values of the masked patches are subsequently predicted, with the mean squared error (MSE) between the reconstructed and true patch values serving as the training objective:
\begin{equation}
\begin{aligned}
    \mathcal{L}_{\mathrm{MPR}} = \sum_{i=1}^{|\mathcal{S}|} \mathbb{E}_{p_{i,j} \in \widetilde{\mathcal{P}}_i} \|\hat{\vp}_{i,j}- \vp_{i,j}\|_2^2, 
\end{aligned}
\label{sce}
\end{equation}
where $\widetilde{\mathcal{P}}_i$ is the set of masked patches for the $i$-th type of spectrum, and $\hat{\vp}_{i,j}$ denotes the reconstruction for the masked patch $\vp_{i,j}$.

\textbf{Contrastive learning between 3D structures and spectra.}
To align spectral and 3D molecular representations, we introduce a contrastive learning objective in addition to the MPR. The 3D embeddings are learned under the supervision of a denoising objective~\cite{Coord,MolSpectra}. Here, the 3D embedding $\vz_x \in \mathbb{R}^d$ and the spectra embedding $\vz_s \in \mathbb{R}^d$ of the same molecule are treated as positive pairs, while all other pairings are considered negative. The contrastive objective is designed to maximize similarity between positive pairs while minimizing similarity with negatives, using the InfoNCE loss~\citep{InfoNCE}:
\begin{equation}
\begin{aligned}
    \mathcal{L}_{\text{Contrast}} = -\frac{1}{2} \mathbb{E}_{p(\vz_x, \vz_s)} [ \log \frac{\exp(f_x(\vz_x, \vz_s))}{\exp(f_x(\vz_x, \vz_s)) + \sum_j \exp(f_x(\vz_x^j, \vz_s))} \\
    + \log \frac{\exp(f_s(\vz_s, \vz_x))}{\exp(f_s(\vz_s, \vz_x)) + \sum_j \exp(f_s(\vz_s^j, \vz_x))} ],
\label{eq:contrast}
\end{aligned}
\end{equation}
where $\vz_x^j$ and $\vz_s^j$ denote negative samples, and $f_x(\vz_x, \vz_s)$ and $f_s(\vz_s, \vz_x)$ are scoring functions, implemented here as the inner product: $f_x(\vz_x, \vz_s) = f_s(\vz_s, \vz_x) = \langle \vz_x, \vz_s \rangle$.

\textbf{Two-stage pre-training pipeline.}
Although spectral datasets are scarce, a wealth of large-scale unlabeled molecular structure datasets exists. By leveraging our proposed contrastive alignment framework between spectra and structures, we transfer knowledge from large-scale pre-training on molecular structures to enhance the learning of SpecFormer.
To fully exploit both spectral and structural information, we adopt a two-stage training protocol: the first stage performs pre-training on a large dataset~\citep{PCQM} containing only 3D structures using the denoising objective, while the second stage utilizes a dataset with available spectra to jointly optimize the overall objective:
\begin{equation}
    \mathcal{L} = \beta_{\text{Denoising}} \mathcal{L}_{\text{Denoising}} + \beta_{\text{MPR}} \mathcal{L}_{\text{MPR}} + \beta_{\text{Contrast}} \mathcal{L}_{\text{Contrast}},
    \label{eq:objective}
\end{equation}
where $\beta_{\text{Denoising}}$, $\beta_{\text{MPR}}$, and $\beta_{\text{Contrast}}$ are weights for each component. In the second stage, SpecFormer is pre-trained exclusively on the QM9S training set to avoid data leakage.
\section{Data availability}
The preprocessed dataset can be accessed at \url{https://huggingface.co/datasets/AzureLeon1/DiffSpectra}.

\section{Code availability}
The source code is open-source and can be accessed at \url{https://github.com/AzureLeon1/DiffSpectra}.
It is available for non-commercial use.
Trained model checkpoints can be found at \url{https://huggingface.co/AzureLeon1/DiffSpectra}.
\section{Acknowledgements}
This work was supported in part by the National Science and Technology Major Project (2023ZD0120901), the Strategic Priority Research Program of Chinese Academy of Sciences (XDA0480102), the Ministry of Education (MOE T1251RES2309 and MOE T2EP20125-0039), the Agency for Science, Technology and Research (ASTAR H25J6a0034), and the Damo Academy through Damo Academy Research Intern Program.

\section{Author Contributions Statement}
L.W., Y.R., and T.X. conceived the initial idea for the project.
L.W. and Z.Z. developed the model.
L.W. processed the dataset and trained the model.
L.W. and Z.Z. carried out the experiments using the trained model.
L.W., T.X., and Z.Z. analyzed the results and created illustrations and data visualizations.
L.W. drafted the initial manuscript.
L.W., Y.R., Z.L, Q.L., Y.Z. participated in the revision of the manuscript.
T.X. and Z.L. provided the instructions on AI modeling.
Z.L., P.W., and Y.Z. provided the instructions on evaluation framework.
The project was supervised by Y.R., D.Z., Q.L., S.W., L.W., Y.Z., with funding provided by Y.R., Q.L., S.W., and Y.Z.

\section{Competing Interests Statement}
The authors declare no competing interests.

\bibliography{main}

\clearpage
\appendix

\captionsetup[figure]{labelformat=empty}
\captionsetup[table]{labelformat=empty}
\renewcommand{\thefigure}{Appendix Figure \arabic{figure}}
\renewcommand{\thetable}{Appendix Table \arabic{table}}

\begin{center}
    \section*{\Large Appendix}
\end{center}
\vspace{2pt}

\section{Derivation of the Reverse Sampling Process}

This appendix derives the reverse sampling formula for our continuous-time diffusion model, as applied to molecular graph generation. Our objective is to obtain an expression for sampling a less-noisy graph state $\cG_s$ from a more-noisy state $\cG_t = (\bH_t, \bA_t, \bX_t)$, given a model prediction of the original clean graph $\hat{\cG}_0 = (\hat{\bH}_0, \hat{\bA}_0, \hat{\bX}_0)$. We assume a continuous time interval $[0, 1]$, with $0 \leq s < t \leq 1$.

To streamline the derivation, we consider a single graph component—either node features $\bH$, edge features $\bA$, or atomic coordinates $\bX$—denoted generically as a vector-valued variable $\bG \in \mathbb{R}^d$. The full derivation extends naturally by applying the result independently to each component.

\subsection{Forward Diffusion Process}

Following the formulation of VDM~\cite{VDM,ProgressiveDistillation,BlurringDiffusion}, the forward diffusion process is defined as:
\begin{equation}
    q(\bG_t \mid \bG_0) = \mathcal{N}(\alpha_t \bG_0, \sigma_t^2 \bI),
    \label{eq:forward}
\end{equation}
where the graph $\bG_t$ at time $t$ is a noisy version of the clean graph $\bG_0$. Equivalently, it can be reparameterized as:
\begin{equation}
    \bG_t = \alpha_t \bG_0 + \sigma_t \bG_\epsilon, \quad \bG_\epsilon \sim \mathcal{N}(\b0, \bI).
\end{equation}
Because $\alpha_t$ decreases monotonically while $\sigma_t$ increases, the information from $\bG_0$ is gradually destroyed as $t$ increase. Assuming the process defined by~\cref{eq:forward} is Markovian, its transition distribution between two intermediate steps $s$ and $t$ with $0 \leq s < t$ is:
\begin{equation}
    q(\bG_t \mid \bG_s) = \mathcal{N}(\alpha_{t|s} \bG_s, \sigma_{t|s}^2 \bI),
\end{equation}
where $\alpha_{t|s} = \frac{\alpha_t}{\alpha_s}$ and $\sigma_{t|s}^2 = \sigma_t^2 - \alpha_{t|s}^2 \sigma_s^2$. A convenient property of this framework is that the time grid can be defined arbitrarily and does not depend on the particular spacing of $s$ and $t$. We set $T = 1$ to denote the final diffusion step, where $q(\bG_T \mid \bG_0) \approx \mathcal{N}(\b0, \bI)$ approximates a standard normal distribution. Unless otherwise specified, time steps are assumed to lie in the unit interval $[0, 1]$.
This formulation describes the distribution of a more-noised state $\bG_t$ conditioned on a less-noised state $\bG_s$, and serves as a key component of the variational framework.

\subsection{Reverse Denoising Process Posterior $q(\bG_s \mid \bG_t, \bG_0)$}

We now derive the posterior distribution $q(\bG_s \mid \bG_t, \bG_0)$, which describes the distribution of an intermediate state $\bG_s$ conditioned on both the noisy future $\bG_t$ and the original clean graph $\bG_0$. This distribution underlies the training of the reverse process.

From standard Bayesian inference for Gaussians, suppose prior $\vz \sim \mathcal{N}(\boldsymbol{\mu}_p, \mathbf{\Sigma}_p)$ and likelihood $\vy \mid \vz \sim \mathcal{N}(\bA \vz, \mathbf{\Sigma}_l)$. Then the posterior $p(\vz \mid \vy)$ is Gaussian with:
\begin{gather}
    \mathbf{\Sigma}_{\text{post}} = \left( \mathbf{\Sigma}_p^{-1} + \bA^\top \mathbf{\Sigma}_l^{-1} \bA \right)^{-1}, \\
    \boldsymbol{\mu}_{\text{post}} = \mathbf{\Sigma}_{\text{post}} \left( \mathbf{\Sigma}_p^{-1} \boldsymbol{\mu}_p + \bA^\top \mathbf{\Sigma}_l^{-1} \vy \right).
\end{gather}
In our case:
\begin{equation}
\begin{aligned}
    \vz = \bG_s, \quad \vy &= \bG_t, \quad \boldsymbol{\mu}_p = \alpha_s \bG_0, \quad \mathbf{\Sigma}_p = \mathbf{\sigma}_s^2 \bI, \\
    \bA &= \alpha_{t|s} \bI, \quad \mathbf{\Sigma}_l = \sigma_{t|s}^2 \bI.
\end{aligned}
\end{equation}
Applying this, we obtain:
\begin{equation}
    q(\bG_s \mid \bG_t, \bG_0) = \mathcal{N}(\boldsymbol{\mu}_Q, \sigma_Q^2 \bI),
\end{equation}
with:
\begin{gather}
    \boldsymbol{\mu}_Q = \sigma_Q^2 \left( \frac{1}{\sigma_s^2} \alpha_s \bG_0 + \frac{\alpha_{t|s}}{\sigma_{t|s}^2} \bG_t \right)
    = \frac{\alpha_{t|s} \sigma_s^2}{\sigma_t^2} \bG_t + \frac{\alpha_s \sigma_{t|s}^2}{\sigma_t^2} \bG_0, \\
    \sigma_Q^2 = \left( \frac{1}{\sigma_s^2} + \frac{\alpha_{t|s}^2}{\sigma_{t|s}^2} \right)^{-1} = \frac{\sigma_s^2 \sigma_{t|s}^2}{\sigma_t^2}.
\end{gather}
The posterior mean is a convex combination of $\bG_t$ and $\bG_0$, with weights determined by their respective noise scales. This provides the optimal denoised estimate of $\bG_s$ and forms the basis of the reverse sampling process.

\subsection{Reverse Denoising Process Approximation}

At inference time, the true $\bG_0$ is unavailable, so we replace it with a model estimate $\hat{\bG}_0 = d_\theta(\bG_t, \tilde{\bG}_0, \log \mathrm{SNR}(t))$. This yields the approximate reverse-time transition:
\begin{equation}
    p(\bG_s \mid \bG_t) \approx \mathcal{N}(\bar{\bG}_s, \sigma_Q^2 \bI),
\end{equation}
where:
\begin{gather}
    \bar{\bG}_s = \frac{\alpha_{t|s} \sigma_s^2}{\sigma_t^2} \bG_t + \frac{\alpha_s \sigma_{t|s}^2}{\sigma_t^2} \hat{\bG}_0, \\
    \sigma_Q^2 = \frac{\sigma_s^2 \sigma_{t|s}^2}{\sigma_t^2}.
\end{gather}
We then sample from this distribution as:
\begin{equation}
    \bG_s = \bar{\bG}_s + \frac{\sigma_s \sigma_{t|s}}{\sigma_t} \bG_\epsilon, \quad \bG_\epsilon \sim \mathcal{N}(\b0, \bI).
\end{equation}
Combining the above expressions, the complete reverse sampling step from $t$ to $s$ becomes:
\begin{equation}
    \bG_s = \frac{\alpha_{t|s} \sigma_s^2}{\sigma_t^2} \bG_t + \frac{\alpha_s \sigma_{t|s}^2}{\sigma_t^2} \hat{\bG}_0 + \frac{\sigma_s \sigma_{t|s}}{\sigma_t} \bG_\epsilon.
\end{equation}
This closed-form expression enables efficient ancestral sampling in the reverse-time generative process.

\subsection{Sampling with Temperature}
To control the stochasticity in the sampling process, we introduce a temperature parameter $\tau > 0$, which scales the noise component during sampling:
\begin{equation}
    \bG_s = \frac{\alpha_{t|s} \sigma_s^2}{\sigma_t^2} \bG_t + \frac{\alpha_s \sigma_{t|s}^2}{\sigma_t^2} \hat{\bG}_0 + \tau \cdot \frac{\sigma_s \sigma_{t|s}}{\sigma_t} \bG_\epsilon.
\end{equation}
The temperature parameter $\tau$ modulates the amount of stochasticity in the sampling process. When $\tau = 1$, the sampling follows the standard formulation. Lower values of $\tau < 1$ reduce the influence of noise, leading to more deterministic and potentially sharper outputs. Conversely, higher values of $\tau > 1$ increase the noise contribution, encouraging greater diversity at the cost of higher stochasticity. This flexibility allows practitioners to trade off between sample fidelity and variability based on downstream objectives.

\section{Datasets}

We conduct experiments on the QM9S dataset~\cite{DetaNet,MolSpectra}, which augments the original QM9~\citep{QM9} molecular dataset with simulated spectra. 
For each molecule, we extract its structure along with the corresponding IR, Raman, and UV-Vis spectra, forming a multi-modal spectral condition for our structure elucidation task.
The dataset contains approximately 127k molecules in total.
Following the dataset split protocol used in previous studies~\cite{EDM,EEGSDE,JODO}, 
we split QM9S into training, validation and test sets, which contain 97K, 17K and 13K samples respectively.

\section{Metrics}

Aiming to generating chemically valid and complete molecules, modeling accurate molecular distributions, and achieving achieve accurate structure elucidation from spectra, we elaborately design a series of evaluation metrics to reflect the generation quality.

Specifically, we introduce two categories of evaluation metrics. The first focuses on general molecular generation quality, emphasizing the validity and chemical soundness of the generated structures. The second targets molecular structure elucidation, evaluating the alignment between the predicted molecular structures—derived from spectral data—and the corresponding ground truth structures.

\subsection{Basic Metrics for Molecular Generation}

In the context of conditional molecule generation, these fundamental evaluation criteria serve as a prerequisite to guarantee that the generated structures are chemically meaningful and sufficiently diverse before evaluating their spectrum-conditioned structural accuracy.

\paragraph{Molecular Structure Validity Evaluation}
The primary consideration is the chemical rationality of the generated molecular structures. The \textbf{Validity} metric ensures that the molecular structures conform to basic chemical rules, including chemically plausible valences (e.g., carbon atoms typically form four covalent bonds), and the overall bonding patterns are chemically reasonable (e.g., avoiding unrealistic bonds between atoms, maintaining aromatic delocalization). 
To further assess the model’s capacity for innovation, we measure the proportions of molecules satisfying \textbf{Validity and Unique (V\&U)} and \textbf{Validity, Unique, and Novelty (V\&U\&N)} criteria. Uniqueness is determined by canonicalizing molecular representations and removing duplicates, thus ensuring diversity among the generated structures. Novelty is evaluated by comparing the generated molecules against the training set, reflecting the model’s ability to produce genuinely novel molecules rather than simply memorizing training data.

\paragraph{Molecular Structure Stability Evaluation}
While structural validity is a necessary condition, it may not fully capture the chemical plausibility of a molecule. We additionally introduce a more stringent indicator of molecular stability. \textbf{Atom Stability} assesses whether each atom in the molecule has achieved an appropriate valence coordination, fully considering the influence of formal charges on the permissible bonding patterns. \textbf{Mol Stability} further requires that all atoms in the entire molecule satisfy the stability criteria, thereby providing a more comprehensive assessment of the chemical soundness of the generated molecules.

\paragraph{Distribution-based Structure Evaluation}
To evaluate the model's ability to learn the data distribution of real molecules, we adopt several distribution-based metrics. The Fréchet ChemNet Distance (\textbf{FCD}) measures the similarity between the distributions of generated molecules and reference molecules in a high-dimensional learned feature space. The \textbf{Similarity to Nearest Neighbor (SNN)} metric assesses the representativeness of the generated set by measuring the Tanimoto similarity between each generated molecule and its nearest neighbors in the test set.
We additionally employ metrics based on molecular structural features: \textbf{Fragment Similarity (Frag)}, derived from BRICS decomposition, evaluates distributional alignment at the functional group level, whereas \textbf{Scaffold Similarity (Scaf)}, based on Bemis–Murcko scaffold analysis, quantifies similarity and diversity at the core structure level. These complementary metrics jointly reflect the model’s ability to capture both local functional groups and global structural patterns, thereby characterizing the complexity of the generated molecular structures.

It is important to note that for all distribution-based metrics, the test set serves as the reference, while the novelty metric uses the training set as the reference to measure whether generated molecules are unseen. This distinction ensures a clear separation between distributional fidelity and generative innovation.

\subsection{Metrics for Molecular Structure Elucidation}\label{sec:metric2}

In the task of molecular structure elucidation, we introduce a set of precise metrics to evaluate the similarity between the ground-truth target structures and the generated candidate molecules under spectrum-conditioned generation. Here, a molecular structure is represented as a graph $\mathcal{G} = (\mathbf{H}, \mathbf{A}, \mathbf{X})$, where $\mathbf{H} \in \mathbb{R}^{N \times d_1}$ denotes node-level attributes such as atom types and charges, $\mathbf{A} \in \mathbb{R}^{N \times N \times d_2}$ encodes pairwise edge features such as bond existence and bond types, and $\mathbf{X} \in \mathbb{R}^{N \times 3}$ corresponds to the 3D coordinates of the atoms.

\paragraph{Top-$K$ Accuracy} 
The \textbf{Top-$K$ Accuracy (Acc@$K$)} quantifies the model’s ability to recover the exact target structure among its top-$K$ generated candidates. For each condition characterized by a set of spectra, the model is sampled $K$ times to generate $K$ molecular structure candidates. If any of the $K$ candidates exactly matches the ground-truth structure, the prediction is considered correct. Formally, the metric is defined as:
\begin{equation}
\operatorname{Acc@}K = \mathbb{E}_{\mathcal{G}} \left[ \mathbbm{1} \left( \exists\, i \in \{1, \ldots, K\}, \; \hat{\mathcal{G}}_i = \mathcal{G} \right) \right],
\end{equation}
where $\mathbbm{1}(\cdot)$ is the indicator function, $\mathcal{G}$ denotes the ground-truth molecular graph, and $\hat{\mathcal{G}}_i$ denotes the $i$-th generated candidate. This metric reflects the probability that the correct structure appears at least once among the top-$K$ model outputs.

\paragraph{Maximum Common Edge Subgraph}
To quantify graph-structural similarity, we adopt the \textbf{Maximum Common Edge Subgraph (MCES)} distance metric:
\begin{equation}
\operatorname{MCES} = \mathbb{E}_{\mathcal{G}} \left[ |\operatorname{E}(\mathcal{G})| + |\operatorname{E}(\hat{\mathcal{G}})| - 2 \cdot \max_{\mathcal{H} \subseteq \mathcal{G},\, \mathcal{H} \subseteq \hat{\mathcal{G}}} |\operatorname{E}(\mathcal{H})| \right],
\end{equation}
where $\mathcal{G}$ and $\hat{\mathcal{G}}$ are the target and predicted molecular graphs respectively, $\operatorname{E}(\mathcal{H})$ denotes the set of edges in the common subgraph $\mathcal{H}$, and $|\operatorname{E}(\mathcal{G})|$ represents the total number of edges in graph $\mathcal{G}$. This distance metric captures the structural dissimilarity between molecular graphs, with lower values indicating greater structural similarity and better recovery of the target molecular structure.

\paragraph{Fingerprint-based Similarity Evaluation}
We adopt several fingerprint-based similarity measures. Denote by $\mathbf{a}_{\mathrm{Morgan}}$ and $\hat{\mathbf{a}}_{\mathrm{Morgan}}$ the binary Morgan fingerprint vectors of the target and predicted molecules, respectively, and by $\mathbf{a}_{\mathrm{MACCS}}$ and $\hat{\mathbf{a}}_{\mathrm{MACCS}}$ their MACCS fingerprints. The \textbf{Tanimoto Similarity over Morgan fingerprints ($\operatorname{TaniSim}_{\mathrm{Morgan}}$)} is defined as:
\begin{equation}
\operatorname{TaniSim}_{\mathrm{Morgan}} = 
\mathbb{E}_{\mathcal{G}} \left[
\frac{ \left| \mathbf{a}_{\mathrm{Morgan}} \land \hat{\mathbf{a}}_{\mathrm{Morgan}} \right| }
     { \left| \mathbf{a}_{\mathrm{Morgan}} \lor \hat{\mathbf{a}}_{\mathrm{Morgan}} \right| }
\right],
\end{equation}
where $\land$ and $\lor$ denote bitwise AND and OR over the fingerprint vectors. This score quantifies the overlap in local structural patterns. The \textbf{Cosine Similarity over Morgan fingerprints ($\operatorname{CosSim}_{\mathrm{Morgan}}$)} is defined as:
\begin{equation}
\operatorname{CosSim}_{\mathrm{Morgan}} = 
\mathbb{E}_{\mathcal{G}} \left[
\frac{ \mathbf{a}_{\mathrm{Morgan}} \cdot \hat{\mathbf{a}}_{\mathrm{Morgan}} }
     { \| \mathbf{a}_{\mathrm{Morgan}} \| \, \| \hat{\mathbf{a}}_{\mathrm{Morgan}} \| }
\right].
\end{equation}
Likewise, the \textbf{Tanimoto Similarity over MACCS fingerprints ($\operatorname{TaniSim}_{\mathrm{MACCS}}$)} is:
\begin{equation}
\operatorname{TaniSim}_{\mathrm{MACCS}} = 
\mathbb{E}_{\mathcal{G}} \left[
\frac{ \left| \mathbf{a}_{\mathrm{MACCS}} \land \hat{\mathbf{a}}_{\mathrm{MACCS}} \right| }
     { \left| \mathbf{a}_{\mathrm{MACCS}} \lor \hat{\mathbf{a}}_{\mathrm{MACCS}} \right| }
\right].
\end{equation}
These fingerprints capture interpretable functional patterns based on a set of predefined substructure keys, complementing the Morgan-based representations.

\paragraph{Fragment-based Similarity Evaluation}
The \textbf{Fragment-based Similarity (Fraggle)} metric decomposes $\mathcal{G}$ into a set of chemically meaningful fragments $\mathcal{F} = \{f_1, f_2, \dots, f_n\}$ by strategies including double acyclic or exocyclic bond cuts with specified fragment size constraints. For each fragment $f_i$, Fraggle computes two Tanimoto similarities over RDKit fingerprints between the target molecule $\mathcal{G}$ and the predicted molecule $\hat{\mathcal{G}}$: one based on the standard RDKit fingerprint over the entire molecule, and another based on a masked RDKit fingerprint where atoms outside $f_i$ with a Tversky similarity below 0.8 are replaced with wildcards. The fragment-level score is then taken as the maximum of these two values:
\begin{equation}
s_i = \max \left( 
\operatorname{TaniSim}_{\mathrm{RDKit}}(\mathbf{a}_{\mathrm{RDKit}}, \hat{\mathbf{a}}_{\mathrm{RDKit}}),
\operatorname{TaniSim}_{\mathrm{RDKit}}(\mathbf{a}_{\mathrm{RDKit,mask}}, \hat{\mathbf{a}}_{\mathrm{RDKit,mask}})
\right).
\end{equation}
The final Fraggle similarity is defined as:
\begin{equation}
\operatorname{FraggleSim} = \mathbb{E}_{\mathcal{G}} \left[
\max_{f_i \in \mathcal{F}} s_i
\right].
\end{equation}
This dual approach captures both global and local matching patterns

\paragraph{Functional Group-based Similarity Evaluation}
To evaluate functional group consistency, we define a set of chemically significant functional groups (e.g., alkane, alcohol, amine, carboxylic acid, etc.) described by SMARTS patterns. For a pair of molecules $\mathcal{G}$ (ground truth) and $\hat{\mathcal{G}}$ (prediction), we extract their respective functional group sets $\operatorname{FG}(\mathcal{G})$ and $\operatorname{FG}(\hat{\mathcal{G}})$. The \textbf{Functional Group-based Similarity (FGSim)} is then computed as:
\begin{equation}
\operatorname{FGSim} = 
\mathbb{E}_{\mathcal{G}} \left[
\frac{ \left| \operatorname{FG}(\mathcal{G}) \cap \operatorname{FG}(\hat{\mathcal{G}}) \right| }
     { \left| \operatorname{FG}(\mathcal{G}) \cup \operatorname{FG}(\hat{\mathcal{G}}) \right| }
\right].
\end{equation}
This interpretable metric is independent of molecular size, computationally efficient via substructure matching, and well-suited for comparative evaluation.

\paragraph{3D Structural Evaluation}

Lastly, to evaluate the accuracy of generated 3D structures, we adopt \textbf{Root Mean Square Deviation (RMSD)} to measure the geometric deviation between the generated 3D structures and ground-truth 3D structures. Prior to RMSD computation, atom-to-atom correspondence must be established between the generated and reference molecules. We employ the Hungarian algorithm to determine the optimal atom mapping by minimizing a weighted distance matrix. The mapping process ensures that each atom in the generated structure is paired with its most similar counterpart in the ground-truth structure based on geometric and chemical criteria.
The RMSD, computed in ångströms (Å), is defined as:
\begin{equation}
\text{RMSD} = \sqrt{\frac{1}{n} \sum_{i=1}^{n} \|\mathbf{X}_i - \hat{\mathbf{X}}_i\|^2}
\end{equation}
where $\mathbf{X}_i$ and $\hat{\mathbf{X}}_i$ represent the 3D coordinates of the $i$-th mapped atom pair in the ground-truth and generated structures, respectively, and $n$ is the number of successfully mapped atoms.

Additionally, we compute the \textbf{Atom Mapping Accuracy (MapAcc)} as:
\begin{equation}
\text{MapAcc} = \frac{1}{n} \sum_{i=1}^{n} \mathbbm{1}[\text{AtomType}(\mathbf{H}_i) = \text{AtomType}(\hat{\mathbf{H}}_i)]
\end{equation}
where $\mathbbm{1}[\cdot]$ is the indicator function, and $\text{AtomType}(\mathbf{H}_i)$ and $\text{AtomType}(\hat{\mathbf{H}}_i)$ denote the atom types of the $i$-th mapped atom pair, providing a complementary measure of structural fidelity that captures both geometric and chemical similarity.

\end{document}